\documentclass[preprint,12pt]{elsarticle}

\usepackage{amssymb}
 
\usepackage{amsmath}

\usepackage[colorinlistoftodos]{todonotes}
\usepackage{array}
\usepackage{float} 
\usepackage{adjustbox} 
\usepackage{titlesec}
\usepackage{arydshln}
\usepackage{hyperref} 
\usepackage{cleveref}
\usepackage{pifont}

\journal{Engineering Applications of Artificial Intelligence}
\usepackage[margin=1.1in,a4paper]{geometry}
\usepackage{booktabs}
\usepackage{caption}
\crefname{section}{Section}{Sections}
\crefname{figure}{Figure}{Figures}
\crefname{table}{Table}{Tables}
\newcommand{\xmark}{\ding{55}}
\usepackage{multirow}  
\usepackage{graphicx}
\begin{document}

\begin{frontmatter}

\title{Ali-AUG: Innovative Approaches to Labeled Data Augmentation using One-Step Diffusion Model}

\author[inst1]{Ali Hamza} 
\author[inst1]{Aizea Lojo}

\affiliation[inst1]{organization={Ikerlan},
            addressline={José María Arizmendiarrieta Pasealekua,2}, 
            city={Mondragon},
            postcode={20500}, 
            state={Gipuzkoa},
            country={Spain}}

\author[inst2,inst3]{Adrian Núñez-Marcos}
\author[inst2,inst3]{Aitziber Atutxa}

\affiliation[inst2]{organization={HiTZ},
            addressline={Manuel Lardizabal Pasealekua, 1}, 
            city={San Sebastián},
            postcode={20018}, 
            state={Gipuzkoa},
            country={Spain}}
            
\affiliation[inst3]{organization={Bilbao School of Engineering (UPV/EHU)},
            addressline={Calle Rafael Moreno “Pitxitxi”, 3}, 
            city={Bilbao},
            postcode={48013}, 
            state={Bizkaia},
            country={Spain}}
            

\begin{abstract}
This paper introduces Ali-AUG, a novel single-step diffusion model for efficient labeled data augmentation in industrial applications. Our method addresses the challenge of limited labeled data by generating synthetic, labeled images with precise feature insertion. Ali-AUG utilizes a stable diffusion architecture enhanced with skip connections and LoRA modules to efficiently integrate masks and images, ensuring accurate feature placement without affecting unrelated image content. Experimental validation across various industrial datasets demonstrates Ali-AUG's superiority in generating high-quality, defect-enhanced images while maintaining rapid single-step inference. By offering precise control over feature insertion and minimizing required training steps, our technique significantly enhances data augmentation capabilities, providing a powerful tool for improving the performance of deep learning models in scenarios with limited labeled data. Ali-AUG is especially useful for use cases like defective product image generation to train AI-based models to improve their ability to detect defects in manufacturing processes. Using different data preparation strategies, including Classification Accuracy Score (CAS) and Naive Augmentation Score (NAS), we show that Ali-AUG improves model performance by 31\% compared to other augmentation methods and by 45\% compared to models without data augmentation. Notably, Ali-AUG reduces training time by 32\% and supports both paired and unpaired datasets, enhancing flexibility in data preparation.

\end{abstract}

\begin{graphicalabstract}

\begin{table}[H]
    \centering
    \renewcommand{\arraystretch}{1.2} 
    \setlength{\tabcolsep}{5pt} 
    \begin{tabular}{>{\centering\arraybackslash}m{0.22\textwidth}>{\centering\arraybackslash}m{0.22\textwidth}>{\centering\arraybackslash}m{0.22\textwidth}>{\centering\arraybackslash}m{0.22\textwidth}}
        \hline
        \textbf{Original Image} & \textbf{Mask} & \textbf{Prompt: "...scratch"} & \textbf{Prompt: "...liquid"} \\
        \includegraphics[width=0.18\textwidth]{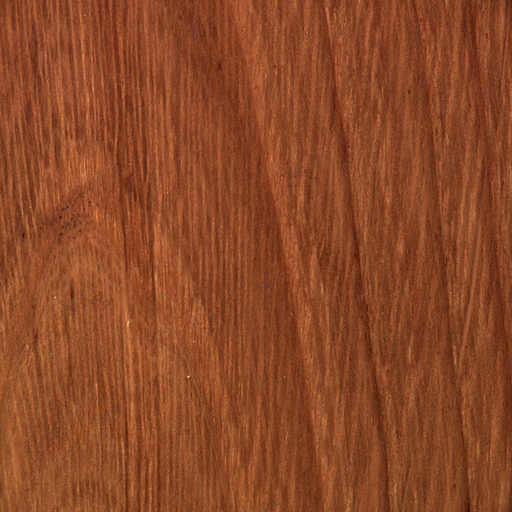} & 
        \includegraphics[width=0.18\textwidth]{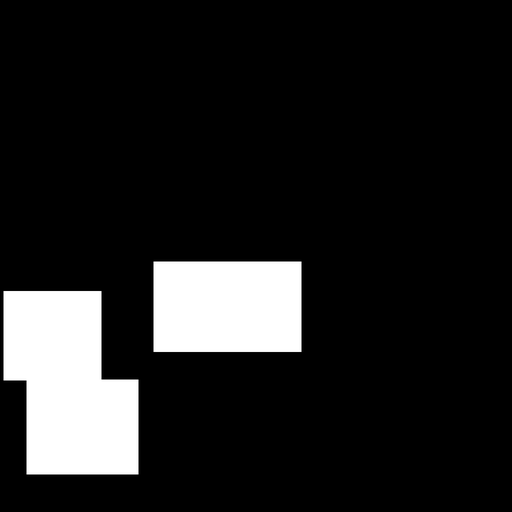} & 
        \includegraphics[width=0.18\textwidth]{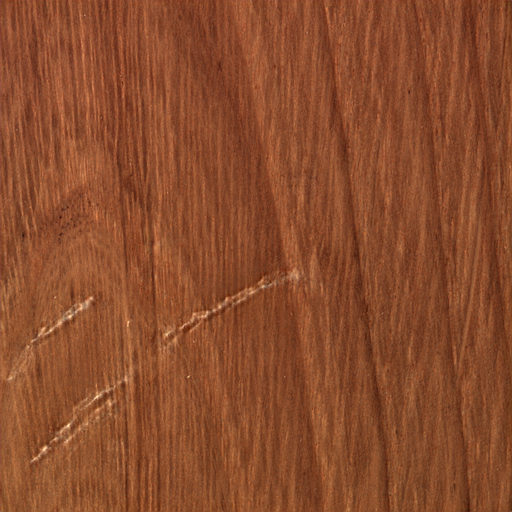} & 
        \includegraphics[width=0.18\textwidth]{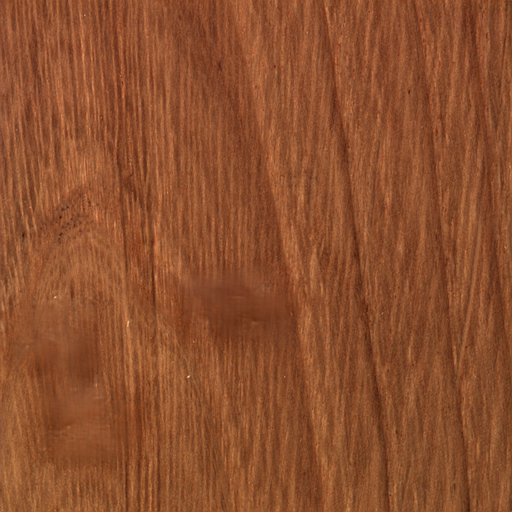} \\
        \hline
        \textbf{Original Image} & \textbf{Mask} & \textbf{Prompt: "...hole"} & \textbf{Prompt: "...liquid"} \\
        "No Input Image" &
        \includegraphics[width=0.18\textwidth]{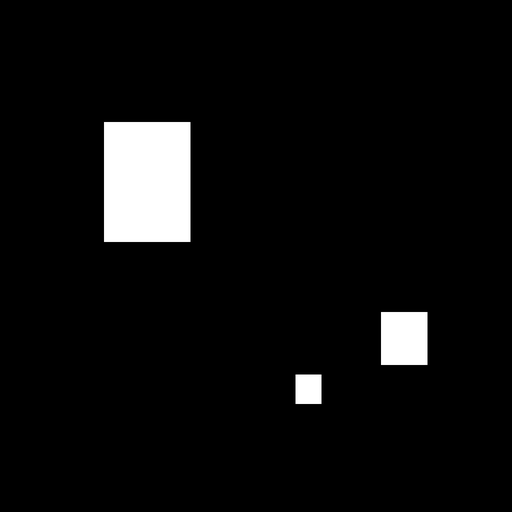} & 
        \includegraphics[width=0.18\textwidth]{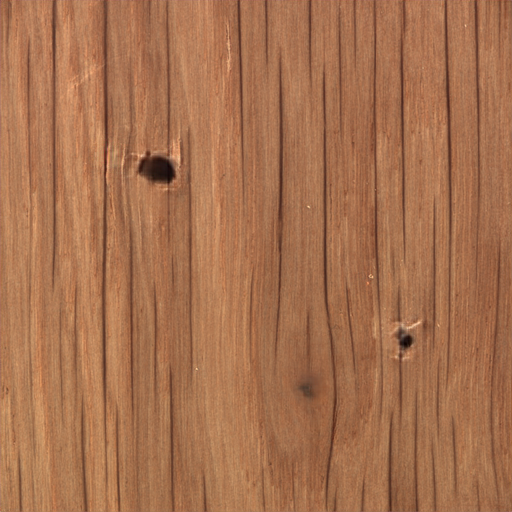} & 
        \includegraphics[width=0.18\textwidth]{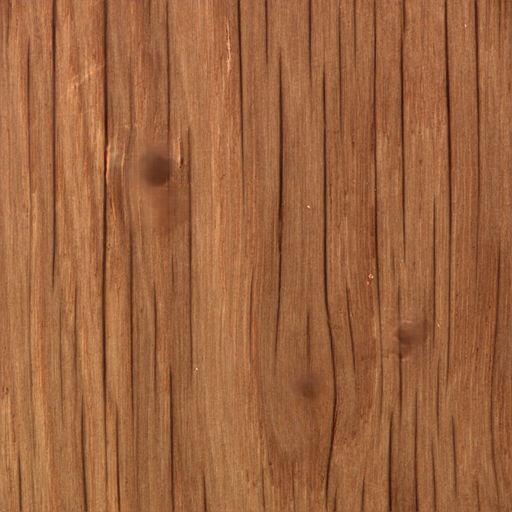} \\
        \hline
        \textbf{Original Image} & \textbf{Mask} & \textbf{Prompt: "...cut"} & \textbf{Prompt: "...color"} \\
        \includegraphics[width=0.18\textwidth]{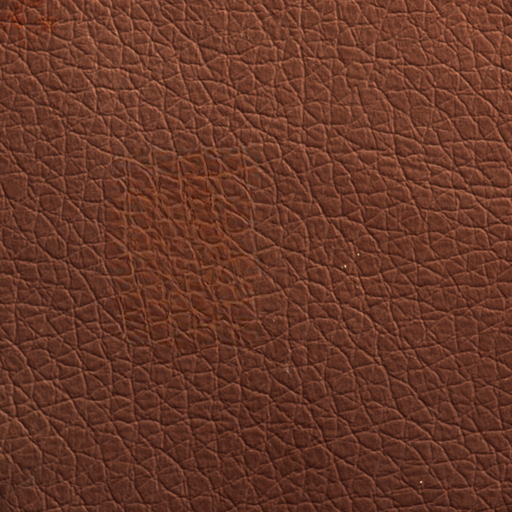} & 
        \includegraphics[width=0.18\textwidth]{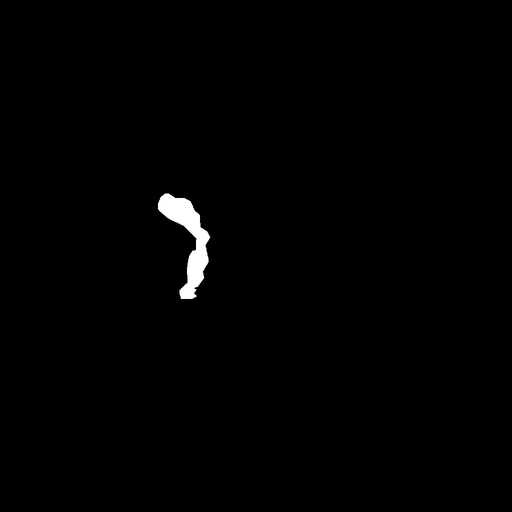} & 
        \includegraphics[width=0.18\textwidth]{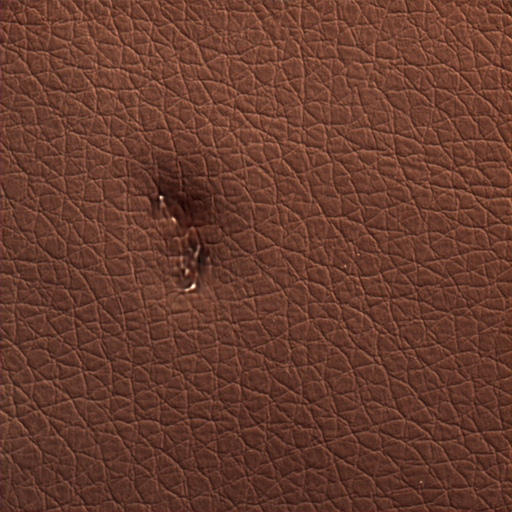} & 
        \includegraphics[width=0.18\textwidth]{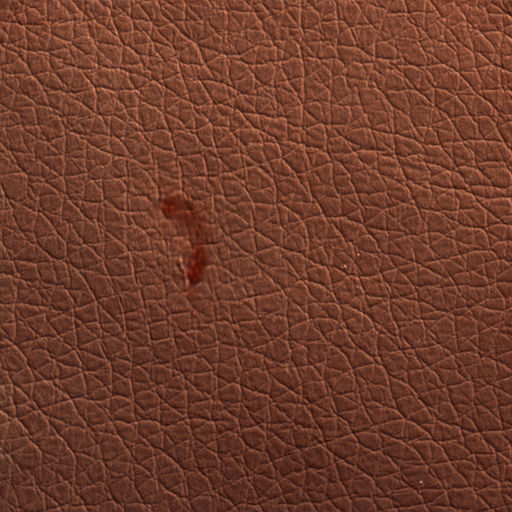} \\
        \hline
        \textbf{Original Image} & \textbf{Mask} & \textbf{Prompt: "...glue"} & \textbf{Prompt: "...color"} \\
        \includegraphics[width=0.18\textwidth]{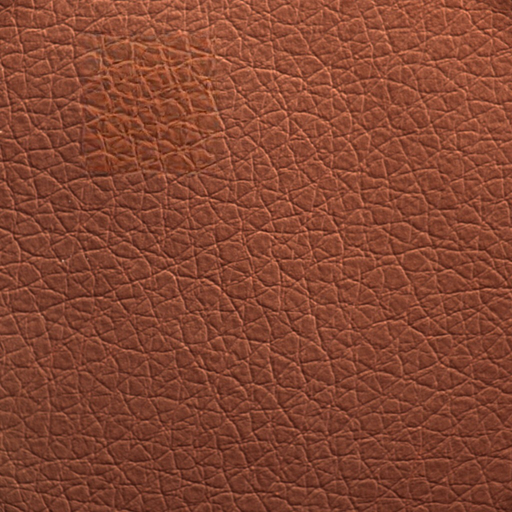} & 
        \includegraphics[width=0.18\textwidth]{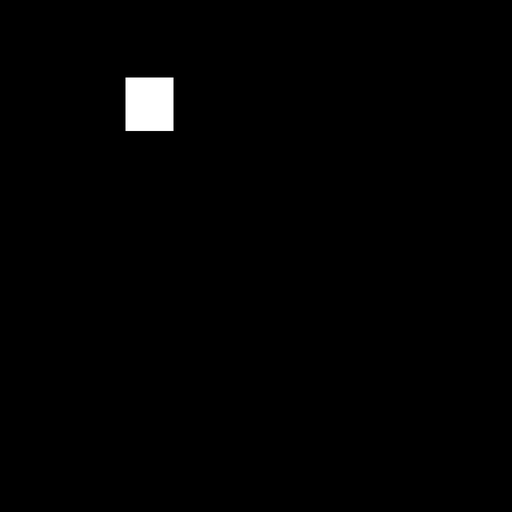} & 
        \includegraphics[width=0.18\textwidth]{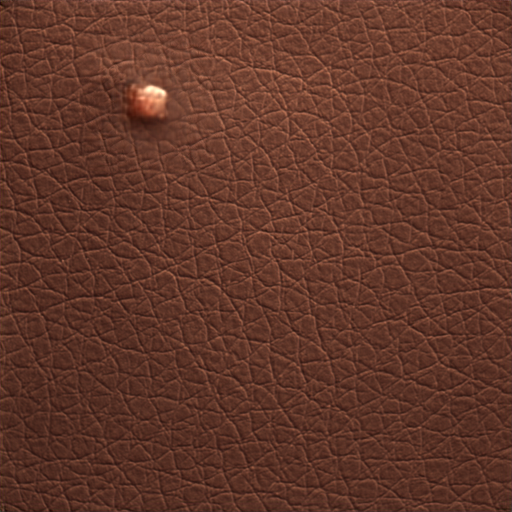} & 
        \includegraphics[width=0.18\textwidth]{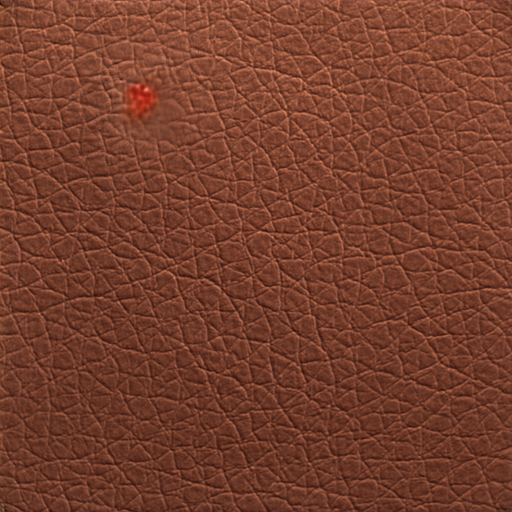} \\
        \hline
    \end{tabular}
    \caption{Paired image generation: The first row shows the complete process with an original image, while the second row demonstrates generation using only a mask input, showing the model's capability to generate images without requiring an original reference.}

    \label{tab:paired_label}
\end{table}

\begin{table}[H]
    \centering
    \renewcommand{\arraystretch}{1.5}
    \setlength{\tabcolsep}{5pt} 
    \begin{tabular}{>{\centering\arraybackslash}m{0.18\textwidth}>{\centering\arraybackslash}m{0.18\textwidth}>{\centering\arraybackslash}m{0.18\textwidth}>{\centering\arraybackslash}m{0.18\textwidth}>{\centering\arraybackslash}m{0.18\textwidth}}
        \hline
        \textbf{Original Image} & \textbf{Mask} & \textbf{Prompt: "...glue strip"} & \textbf{Prompt: "...oil"} & \textbf{Prompt: "...gray stroke"} \\
        \includegraphics[width=0.16\textwidth]{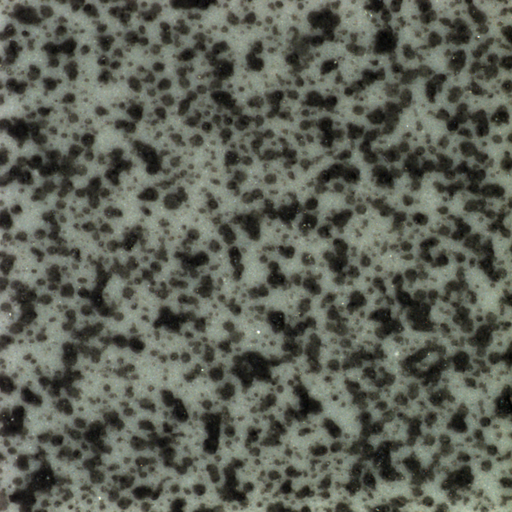} & 
        \includegraphics[width=0.16\textwidth]{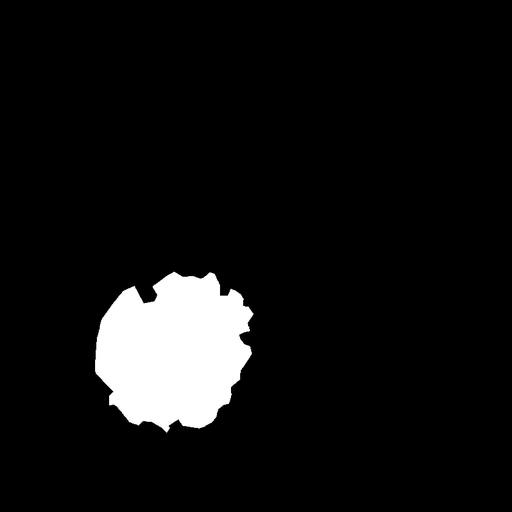} & 
        \includegraphics[width=0.16\textwidth]{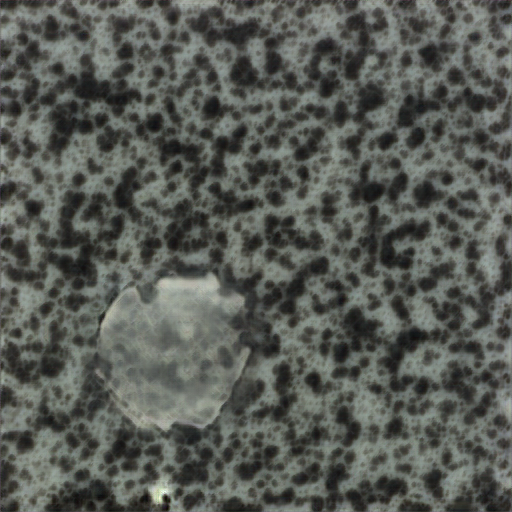} & 
        \includegraphics[width=0.16\textwidth]{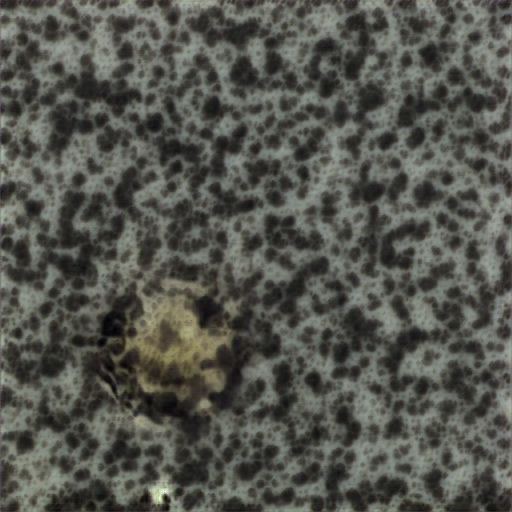} &
        \includegraphics[width=0.16\textwidth]{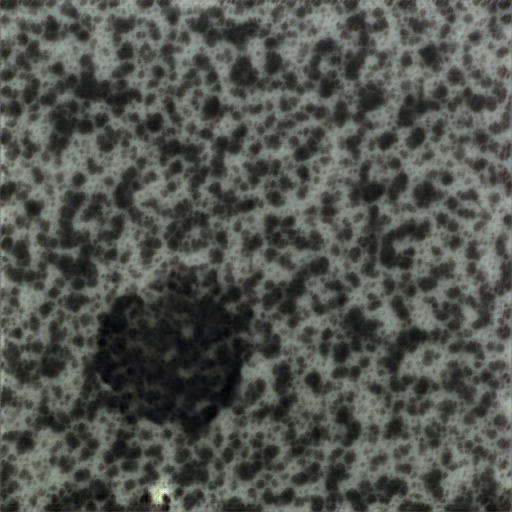} \\
        \hline
    \end{tabular}
    \caption{Unpaired: Image generation process from masks and prompts}
    \label{tab:unpaired_label}
\end{table}

\end{graphicalabstract}

\begin{highlights}
\item Ali-AUG: A novel single-step diffusion model for efficient labeled data augmentation
\item Precise feature insertion using masks and prompts without affecting unrelated image content
\item Support for both paired and unpaired training settings, offering flexibility in data preparation
\item Significant reduction in training time and computational resources compared to existing models
\item Improved performance in downstream tasks, particularly in scenarios with limited labeled data
\end{highlights}

\begin{keyword}
 Data Augmentation \sep Single-Step Diffusion Models \sep Labeled Data \sep Training Time Reduction \sep Industrial Applications \sep Defective Product Image Generation
\end{keyword}

\end{frontmatter}



\newpage
\section{Introduction}  \label{sec:introduction}

In industrial applications, cameras, sensors, and other monitoring devices are prevalent for product quality control and inspection. Typically, the data from these cameras and sensors are used to train traditional AI models such as decision trees. However, most of these AI models rely on supervised algorithms, which require human-annotated instances, where each instance is assigned a specific label or class to be learned. The automatic identification of defective products is a common example, where the AI model training dataset must include enough labeled instances categorized as either 'good' or 'defective'. 
The generation of these datasets presents two significant challenges in the industrial context. First, the scarcity of defective instances results in models struggling to detect defective products accurately. This issue is critical across various fields, such as fault detection in industrial components, early disease identification in medical imaging, and structural health monitoring of infrastructure. Second, personnel and time constraints, especially in small or mid-size enterprises. Generating the necessary datasets presents significant obstacles beyond the scarcity of defective data. The labeling process is labor-intensive and time-consuming, further complicating the task.

In recent years, the AI field has shifted from traditional machine learning models to deep learning approaches, which have shown exceptional performance in computer vision (CV) tasks such as image classification, object detection, image segmentation, and medical image categorization. However, despite their success, deep learning models require bigger data amounts than traditional AI models to achieve high accuracy, making them heavily dependent on large, well-labeled datasets. These models effectively learn different features of an image by applying convolution operations, with initial layers focusing on low-level features (e.g., edges, lines) and deeper layers capturing more complex and structured features. Nonetheless, their performance decreases drastically in low-data regimes and often suffers from overfitting, where the model performs well on training data but poorly on unseen test data. This problem is exacerbated, particularly in industry or clinical domains, by the scarcity of large, labeled datasets due to privacy concerns and the time-consuming nature of human labeling tasks \cite{kumar2021binary,kumar2023imagedataaugmentationapproaches,yun2019cutmixregularizationstrategytrain}.

To address this issue, the generation of synthetic datasets has recently gained popularity. Models such as Generative Adversarial Networks (GANs) \cite{liu2021pdganprobabilisticdiversegan,zheng2022cmganimageinpaintingcascaded} and autoencoders \cite{peng2021generating,zheng2019pluralisticimagecompletion} are capable of generating realistic images; however, they still rely on extensive images to generate more \cite{Shao2019GenerativeAN}. 

This work proposes a novel system for synthetically generating labeled datasets, with the aim of producing images conditioned on specific constraints, such as the presence of a particular defect. Conditioned generation allows us to have better control of the generation process. There are various methods to condition generated images. Initially, Generative Adversarial Networks (GANs) were used for this purpose, but the results were often inaccurate. Later, diffusion models \cite{huang2024diffusionmodelbasedimageediting} were developed offering improved conditioning capabilities. For instance, we can condition a model by providing it with a textual prompt and/or a mask. A mask is a binary or continuous map that specifies which parts of the image should be altered or remain unchanged, which ensures that the generated image adheres precisely to the shape and details specified by this mask. Although some diffusion models like \cite{zhangAddingConditionalControl2023,juBrushNetPlugandPlayImage2024, Wu_2023_ICCV, wassermanPaintInpaintLearning2024,lugmayrRePaintInpaintingUsing2022,corneanuLatentPaintImageInpainting2024} are capable of conditioning images effectively using masks, they still face limitations related to the extensive training requirements and slow inference times because they incorporate one or more additional U-Nets specifically designed to process conditioning inputs. U-Nets are specific instances of a neural network that follow the U-Net architecture and can capture hierarchical features at multiple scales. These additional U-Nets increase computational complexity, resulting in longer training times. 

Examples are ControlNet \cite{zhang2023addingconditionalcontroltexttoimage}, and BrushNet \cite{juBrushNetPlugandPlayImage2024}. They incorporate the mask into a simplified image representation called the latent space, derived from encoding the image using a Variational Autoencoder (VAE). This approach arises because conditioning images directly at full resolution would be computationally expensive. By encoding the image into a lower-dimensional space (often reduced by a factor of 8), these models reduce computational complexity, as the conditioning is applied to this compressed representation rather than the original image. However, this comes at the cost of precision, especially when making minor alterations. Small or fine-grained details are harder to manipulate accurately, as the compression in the latent space loses some of the detailed information. While this method can simplify training and reduce the need for large datasets, it still results in long training times. ControlNet, despite enabling structural conditioning with masks, struggles with capturing fine details because of the limitations imposed by the compressed representation. Similarly, BrushNet, which is designed for inpainting, sometimes fails to place characteristics in the exact location, likely due to operating within the latent space rather than the full-resolution image.

There is also an img2img-turbo \cite{parmarOneStepImageTranslation2024} model (see Figure \ref{fig:parmar_model}) that reduces inference to a single step but, like the aforementioned models, still takes a long time to train. The purpose of the img2img-turbo model is to generate images from a sketch or canny inputs rather than perform inpainting.

Within this framework, we introduce a novel approach for producing synthetic, labeled images and tailoring a variety of images constrained by specific conditions in the form of masks and textual prompts focused on industrial applications. Our method addresses inefficiencies in existing models by offering precise control over the insertion of specific features, such as manufacturing defects or malformations, using a prompt to define the characteristics of the alteration and a mask to specify its location. Furthermore, our approach can be efficiently adapted to specific datasets, including complex datasets that go beyond binary classification. This allows for the generation of datasets for more intricate tasks such as multi-class classification, object detection with multiple defects, and even unseen types of defects not present in the original data. By enabling the creation of such complex datasets, our method supports the development of models capable of handling diverse and nuanced real-world scenarios. Our model can be trained on a single GPU with 24GB of memory in just a few hundred epochs. It operates on both original-defective paired image datasets and unpaired original image datasets, as detailed below.

 \subsection{Paired Dataset Example}
In the paired dataset example shown in Table \ref{tab:paired_example}, each pair contains the image of an object and its defective version. This allows for direct comparison and assessment of the model's ability to introduce defects while maintaining the integrity of the rest of the image content. The mask guides the specific location of modifications, ensuring that only the designated areas are altered, and the prompt controls the modification type. For instance, a prompt like "add scratch" or "add hole" will introduce these specific characteristics to the image using the provided mask, allowing for precise and controlled defect generation.

\begin{table}[H]
\centering
\begin{tabular}{cccc}
\hline
Mask  & Input Image  & Prompt  & Ground truth  \\ 
\hline
\includegraphics[width=0.2\textwidth]{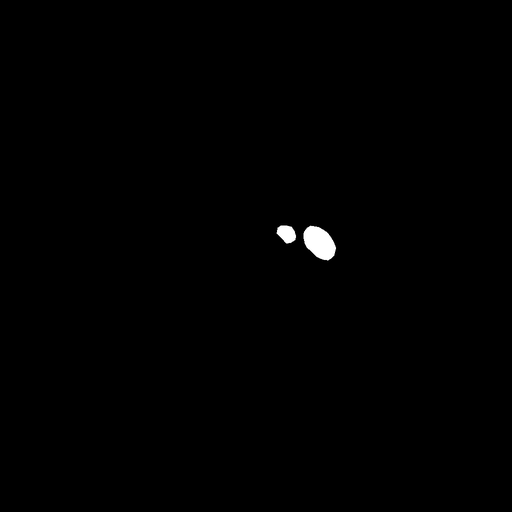} & 
\includegraphics[width=0.2\textwidth]{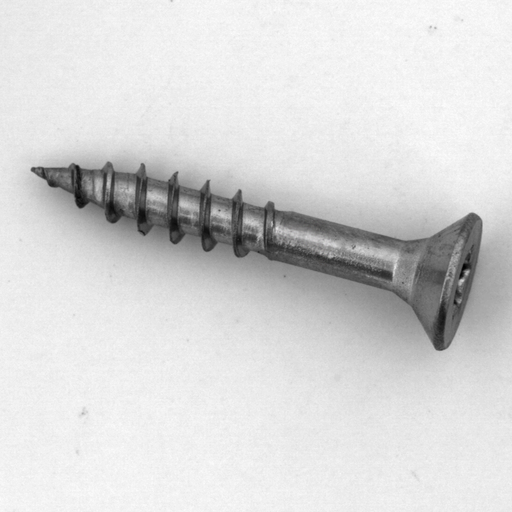} & 
\multirow{-6}{*}{\centering \texttt{"add scratch"}} & 
\includegraphics[width=0.2\textwidth]{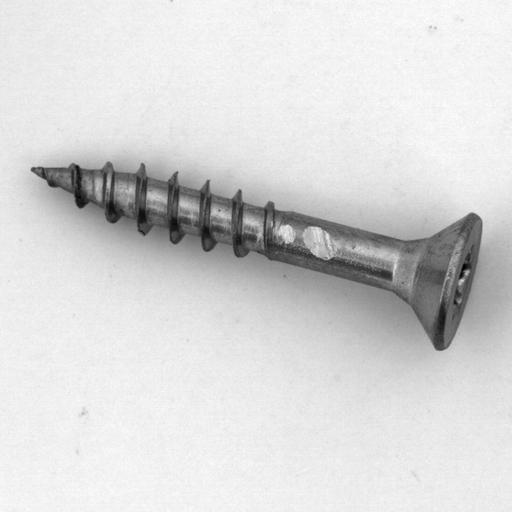} \\
\hline
\end{tabular}
\caption{Example of PAIRED DATASET. In a paired dataset, each entry consists of an original input image and its corresponding defective version, which is derived from the original.}
\label{tab:paired_example}
\end{table}

 \subsection{Unpaired Dataset Example}

In the unpaired dataset example shown in Table \ref{tab:unpaired_example}, the input image is a defect-free image that has similar visual characteristics (i.e., form, color, texture), but it is not the same object. This scenario is more common in real-world applications where paired data might not be readily available. By using unpaired data, our approach demonstrates flexibility and robustness in generating defects that adhere to the domain characteristics without relying on identical image pairs.

\begin{table}[H]
\centering
\begin{tabular}{ccccc}
\hline
Mask  & Input Image  & Prompt  & Image With Defect \\
\hline
\includegraphics[width=0.2\textwidth]{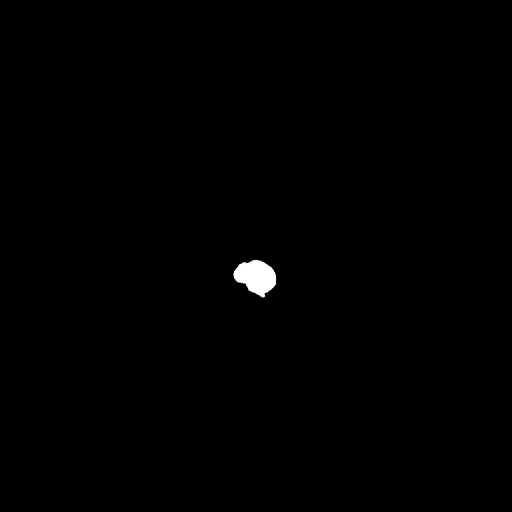} &
\includegraphics[width=0.2\textwidth]{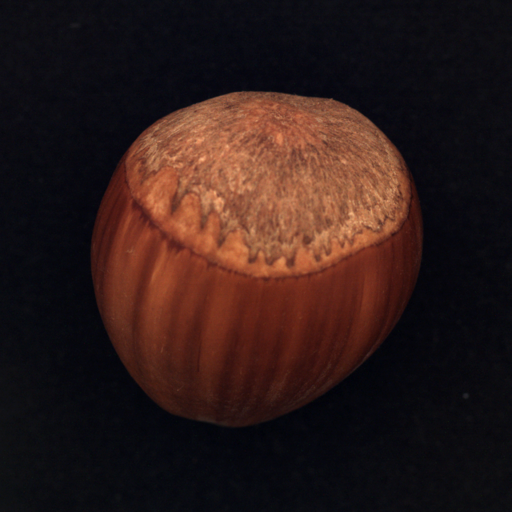} &
\multirow{-6}{*}{\centering \texttt{"add hole"}} &
\includegraphics[width=0.2\textwidth]{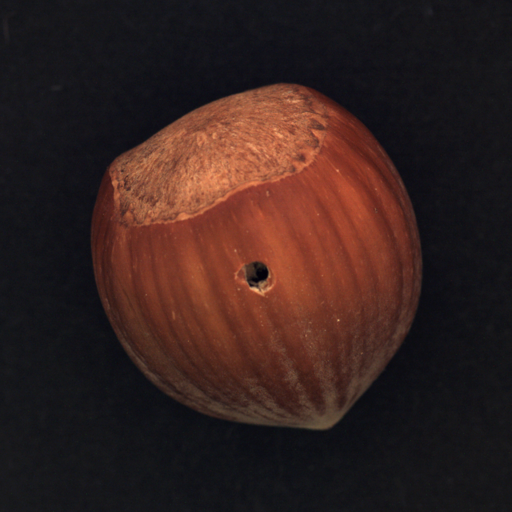} \\
\hline
\end{tabular}
\caption{Example of UNPAIRED DATASET. In an unpaired dataset, the input image is a defect-free image from the same domain, but it is not the same as the image with the defect. This is the most common scenario.}
\label{tab:unpaired_example}
\end{table}

Our approach has the following key innovations:
\begin{itemize}

    \item \textbf{Auto-Labeled Data Augmentation}: Our approach facilitates the creation of synthetic, labeled datasets using few samples. This is particularly valuable for industrial applications where labeled data is scarce or expensive to obtain. Ali-AUG enables data generation with defects or anomalies, which are rarer and more challenging to acquire.

    \item \textbf{Single-Step Conditioning}: We propose a single-step conditioning mechanism that significantly reduces inference time while maintaining high-quality outputs.
    
    \item \textbf{Low-Rank Adaptation (LoRA) Modules \cite{hu2021loralowrankadaptationlarge}}: We enhance the stability and efficiency of our model by using LoRA modules. These modules adapt network weights to the new conditions provided by the mask, reducing the number of parameters that need training and improving the overall performance.
    
    \item \textbf{Support for Paired and Unpaired Training Settings with the Same Model}: Our \textit{Ali-AUG} method is versatile, supporting both paired (see Table \ref{tab:paired_example}) and unpaired (see Table \ref{tab:unpaired_example}) training settings. This flexibility can be applied in various industrial scenarios, even when paired data is scarce. To enable replication of our work, we will publish both the paired and unpaired datasets. 

    \item \textbf{Complex Task Dataset Generation}: Our system extends beyond binary classification by enabling the generation of datasets for more complex tasks, such as multiclass classification and object detection involving multiple defects. This capability allows for creating datasets with various types of defects, including those not present in the original data, thereby enhancing the model's robustness and generalization.

\end{itemize}

\textit{Ali-AUG} delivers visually appealing outcomes while significantly reducing inference times due to its single-step denoising process. In comparison, our \textit{Ali-AUG} model matches the performance of recent developments like \cite{lugmayrRePaintInpaintingUsing2022,xie2023smartbrush,xie2023dreaminpaintertextguidedsubjectdrivenimage},  but it benefits from single-step inference, requires fewer training steps, and uses a very low number of parameters for training using LoRA adapters. By generating labeled synthetic images, our \textit{Ali-AUG} approach efficiently meets the demands for training and inference in computer vision applications, which require accurate and diverse datasets. This strategy can potentially revolutionize data preparation in machine learning, particularly in fields where acquiring large, labeled datasets is difficult or costly. \textbf{We will demonstrate this in the experiments section (\ref{sec:experiments})} by first conducting experiments on a private internal industrial dataset, which cannot be publicly released, followed by experiments on public datasets to ensure the reproducibility of our work. It is important to note that while the approach will be fully described, the underlying code will not be publicly shared in this paper.

\section{Related Work} \label{sec:related_work}
Data augmentation is a common task in computer vision. Basic data augmentation is based on image transformations such as rotation, flipping, changing contrast, and zooming in, among others. Most of these techniques are easy to implement and are meaningful to apply under the assumption that the existing data closely follows the actual data distribution \cite{yang2022image}. However, some basic image manipulation methods, such as translation and rotation, suffer from the padding effect. This effect occurs when parts of the image are moved out of the boundary and get lost after the operation, necessitating interpolation methods to fill in the blank areas.

\begin{figure}[H]
    \centering
    \includegraphics[width=0.95\textwidth]{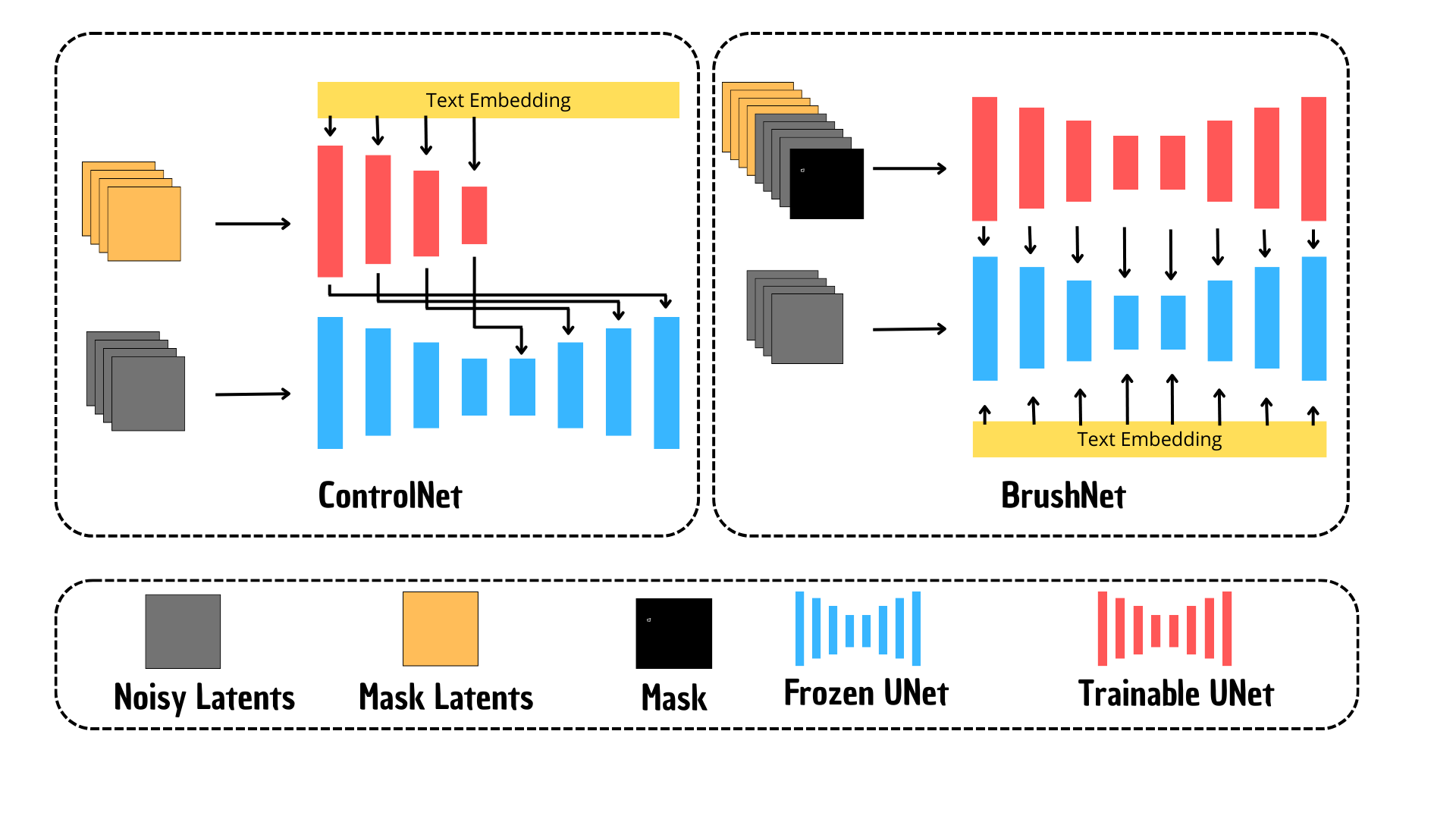}
    \caption{BrushNet \cite{juBrushNetPlugandPlayImage2024} and ControlNet \cite{zhangAddingConditionalControl2023} models architecture. }
    \label{fig:brushnet_controlnet}
\end{figure}

Previous methods based on traditional techniques, such as autoencoders (AEs) and GANs, have achieved satisfactory results, but conditioning GAN networks is challenging \cite{ren2019mask}. Accurate conditioning is essential when using a mask to guide the image generation process, as the mask serves as the label for the generated image. Recently, diffusion models \cite{ho2020denoisingdiffusionprobabilisticmodels,song2022denoisingdiffusionimplicitmodels} have gained popularity due to their ability to generate high-quality images and the control they offer over the image generation process \cite{zhangAddingConditionalControl2023,openai_dalle3,midjourney_home}. Since applying the condition directly to the images is not feasible, diffusion models typically reduce the image to a lower-dimensional latent space, condition the image in this space, and then decode the image from the latent space \cite{rombachHighResolutionImageSynthesis2022,zhangAddingConditionalControl2023,juBrushNetPlugandPlayImage2024}. Most of these models are designed to be plug-and-play with different diffusion models adding an overhead to the model. For example, as mentioned in the introduction, BrushNet and ControlNet (see Figure \ref{fig:brushnet_controlnet}) add an extra U-Net to the existing U-Net to apply mask conditioning, making the training process expensive \cite{zhangAddingConditionalControl2023}. These models intend to guide the generation process, whereas we aim to use the mask itself as a label, ensuring precise alignment between the guidance and the generated output.

\subsection{Data Augmentation with diffusion models.}

Recent advances have shown the potential of diffusion models for data augmentation. Stable Diffusion models are a specific implementation of the diffusion models that utilize a U-Net-based architecture.  Stable Diffusion often incorporates conditioned guidance mechanisms, such as class labels, masks, or text prompts, to control the image generation process. For instance, the DIFFUSEMIX \cite{islamDIFFUSEMIXLabelPreservingData} generates diverse samples based on tailored conditional prompts, which are textual descriptions of the intended conditions. This method allows to effectively combine original and generated images to create hybrid images, enhancing visual diversity while preserving key semantics. DIFFUSEMIX overcomes the drawbacks of using only generated images, which can sometimes produce images that don't look realistic or fail to match the expected labels accurately. By blending generated content with original images, DIFFUSEMIX ensures that the augmented images maintain a natural appearance and retain the correct semantic information, reducing the risk of misleading the training process \cite{islamDIFFUSEMIXLabelPreservingData}. 
 

The DA-Fusion \cite{trabuccoEffectiveDataAugmentation2023} method proposes a flexible data augmentation strategy using text-to-image diffusion models. This approach edits images to change their semantics using an off-the-shelf diffusion model and generalizes to novel visual concepts from a few labeled examples. Evaluated on few-shot image classification tasks and a real-world weed recognition task, DA-Fusion demonstrates improved accuracy and diversity in the augmented data \cite{trabuccoEffectiveDataAugmentation2023}. While these methods significantly improve visual diversity and data augmentation efficiency, the labeling challenge remains.


\subsection{Advancements in Image Editing.}


Unlike image generation, which focuses on creating new images from scratch, and image restoration and enhancement, which aim at repairing and improving the quality of degraded images, image editing involves modifying existing images in terms of appearance, structure, or content. This includes tasks like adding objects, replacing backgrounds, and altering textures \cite{huang2024diffusionmodelbasedimageediting}.

Image editing can be broadly categorized into three types:

\begin{itemize}
    \item \textbf{Semantic Editing:} This involves altering the content and narrative of an image, affecting the depicted scene’s story, context, or thematic elements. Tasks include object addition, object removal, object replacement, background change, and emotional expression modification.
    \item \textbf{Stylistic Editing:} This focuses on enhancing or transforming the visual style and aesthetic elements of an image without altering its narrative content. Tasks include color change, texture change, and overall style change, encompassing both artistic and realistic styles.
    \item \textbf{Structural Editing:} This pertains to changes in the spatial arrangement, positioning, viewpoints, and characteristics of elements within an image, emphasizing the organization and presentation of objects within the scene. Tasks include object movement, object size, and shape change, object action and pose change, and perspective/viewpoint change.
\end{itemize}

Text-guided image editing \cite{huang2024diffusionmodelbasedimageediting}, leveraging human-issued textual commands, initially employed GAN frameworks \cite{reed2016generativeadversarialtextimage}, which often produced unrealistic results. More recently, large-scale trained diffusion models have effectively performed image transformations by controlling cross-modal attention maps \cite{kim2017learning,taigman2016unsupervised}. Sophisticated models like T2I-Adapter \cite{mouT2IAdapterLearningAdapters2023}, BrushNet \cite{juBrushNetPlugandPlayImage2024}, and ControlNet \cite{zhangAddingConditionalControl2023} have introduced fine-tuning techniques using gated transformers or zero-convolution layers, improving the editing precision, although requiring extensive training time and numerous training pairs.

These models \cite{avrahami2022blended,xie2023smartbrush,rombach2022highresolutionimagesynthesislatent,manukyan2024hdpainterhighresolutionpromptfaithfultextguided,chen2024virtualmodelgeneratingobjectidretentivehumanobject,wang2023imagen,zhang2023addingconditionalcontroltexttoimage,juBrushNetPlugandPlayImage2024} are not only costly to train but also exhibit slow performance both during training and inference as shown in Table \ref{tab:comparison}. Additionally, some models inadvertently alter non-targeted parts of the image, and many do not have open-source code, which can be undesirable \cite{huang2024diffusionmodelbasedimageediting}. 

Innovations such as InstructPix2Pix \cite{brooksInstructPix2PixLearningFollow2023} utilize GPT-3 to generate prompts and pair them with Stable Diffusion for image creation, enabling precise edits based on textual instructions. MGIE \cite{fuGuidingInstructionbasedImage2024} further enhances this by integrating a multimodal large language model, although sometimes inadvertently altering non-targeted image areas.


\subsection{One-Step diffusion models.}

To improve the efficiency of diffusion model inference, recent research has concentrated on minimizing the number of sampling steps by employing rapid Ordinary Differential Equation (ODE) solvers \cite{lu2022dpm} or by transforming slow, multi-step teacher models into swift, few-step student models. Direct regression from noise to images typically results in blurry outputs \cite{parmarOneStepImageTranslation2024}. To address this, various distillation techniques incorporate methods such as consistency model training \cite{meng2023distillation,salimans2022progressive}, adversarial learning \cite{sauer2023adversarial}, and variational score distillation \cite{wang2023prolificdreamer}. Parmar et al. \cite{parmarOneStepImageTranslation2024} introduced a stable diffusion model, fine-tuned using LoRA adapters \cite{hu2021loralowrankadaptationlarge} and a single-step scheduler. Their model, despite the slow learning process, was designed to generate an image based on a given mask and prompt. Unlike these works, which focus on one-step text-to-image synthesis and mask-to-image, we present, to our knowledge, the first one-step image editing model for data augmentation that can be trained on a single GPU with 24 GB.

\begin{figure}[H]
    \centering
    \includegraphics[width=0.95\textwidth]{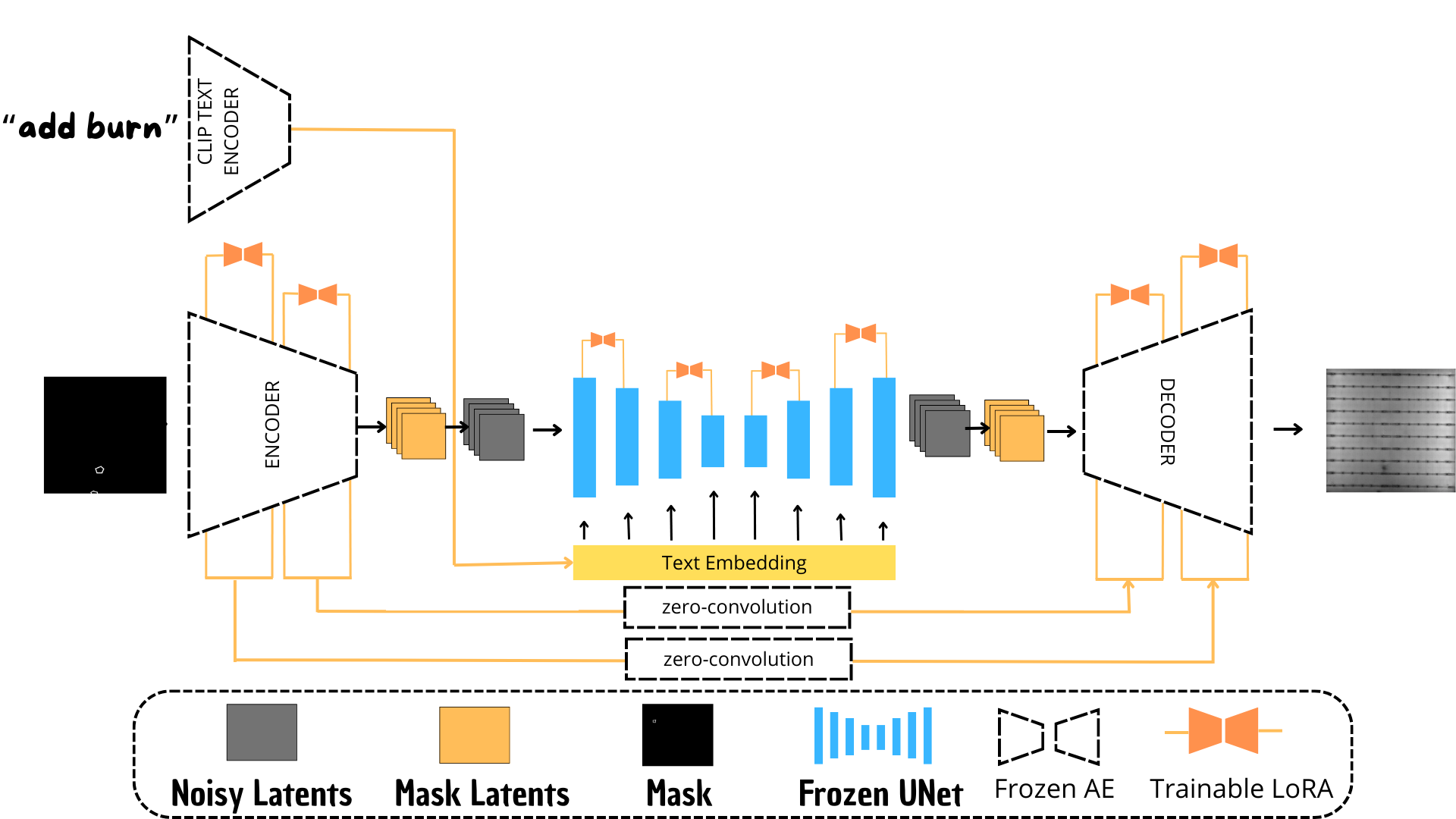}
    \caption{Architecture of the One-Step Image Translation Model. This architecture enables the translation of an input image \( x \) to an output image \( y \) while preserving the structure of the input scene \cite{parmarOneStepImageTranslation2024}, using prompts to guide the creation of images from sketches or canny inputs.}

    \label{fig:parmar_model}
\end{figure}

\subsection{Our Contribution.}

The \textit{Ali-AUG} method advances previous developments by introducing the first one-step conditional model that combines text, input images, and conditioning images to refine the process of text-guided image editing. Unlike studies focused solely on single-step text-to-image synthesis or mask-to-image transformations, our approach uses an initial image for modification, a mask to delineate the specific area for alteration, and a text prompt to guide the process. This method ensures precise and efficient image synthesis by modifying only the targeted area as specified by the text instruction, thereby minimizing undesired changes. Not only does this offer enhanced control, but it also adapts to various image types, significantly broadening the applicability of text and mask conditioning in text-guided editing. The distinct advantages of \textit{Ali-AUG} in terms of plug-and-play capability, training time, precise shape control, inference steps, and training parameters are highlighted in Table \ref{tab:comparison}.

\begin{table}[H]
    \centering
    \small
    \resizebox{\textwidth}{!}{
        \begin{tabular}{lccccc}
            \toprule
            Model  & Plug-and-Play & Training Time & Precise Shape & Inf Steps & Training Params \\
            \midrule
            Blended Diffusion \cite{avrahami2022blended} & \checkmark  & high & \xmark & $>1$  &  mid\\ \hdashline
            SmartBrush \cite{xie2023smartbrush}  &  & high & \checkmark & $>1$  & high \\ \hdashline
            SD Inpainting \cite{rombach2022highresolutionimagesynthesislatent}   &  & mid  &  &$>1$  & high\\ \hdashline
            HD-Painter \cite{manukyan2024hdpainterhighresolutionpromptfaithfultextguided} & \xmark  & high & & $>1$ & high\\ \hdashline
            ReplaccAnything \cite{chen2024virtualmodelgeneratingobjectidretentivehumanobject}  &  & high & \checkmark & $>1$  & high \\ \hdashline
            Imagen \cite{wang2023imagen}  &  & high & \checkmark & $>1$ & high \\ \hdashline
            ControlNet-Inpainting \cite{zhang2023addingconditionalcontroltexttoimage} & \checkmark   & high & \checkmark & $>1$ & mid\\ \hdashline
            BrushNet \cite{juBrushNetPlugandPlayImage2024} & \checkmark & mid & \checkmark & $>1$ & mid \\ 

            \midrule
            \textit{Ali-AUG} & \checkmark & low & \checkmark & 1 & low \\
            \bottomrule
        \end{tabular}
    }

    \caption{Comparison of different models based on various features. If a cell in the row for a model is empty, it means the feature is not mentioned in the corresponding paper. Plug-and-Play indicates whether the model can be easily integrated with other diffusion models. Training Time is the duration needed to train the model, classified as high, mid, or low. Precise Shape denotes whether the model can generate shapes that accurately match the mask. Inf Steps indicates the number of steps needed for inference. Training parameters refers to the number of parameters that need to be trained. Unlike models focusing solely on single-step text-to-image synthesis or mask-to-image transformations, the \textit{Ali-AUG} method utilizes an initial image, a mask to define the specific area for alteration, and a text prompt for precise and efficient text-guided image editing. This comparison highlights the distinct advantages of \textit{Ali-AUG} regarding plug-and-play capability, training time, precise shape control, inference steps, and training parameters.}

    \label{tab:comparison}
\end{table}

\section{Method} \label{sec:method}

\begin{figure}[H]
    \centering
    \includegraphics[width=0.95\textwidth]{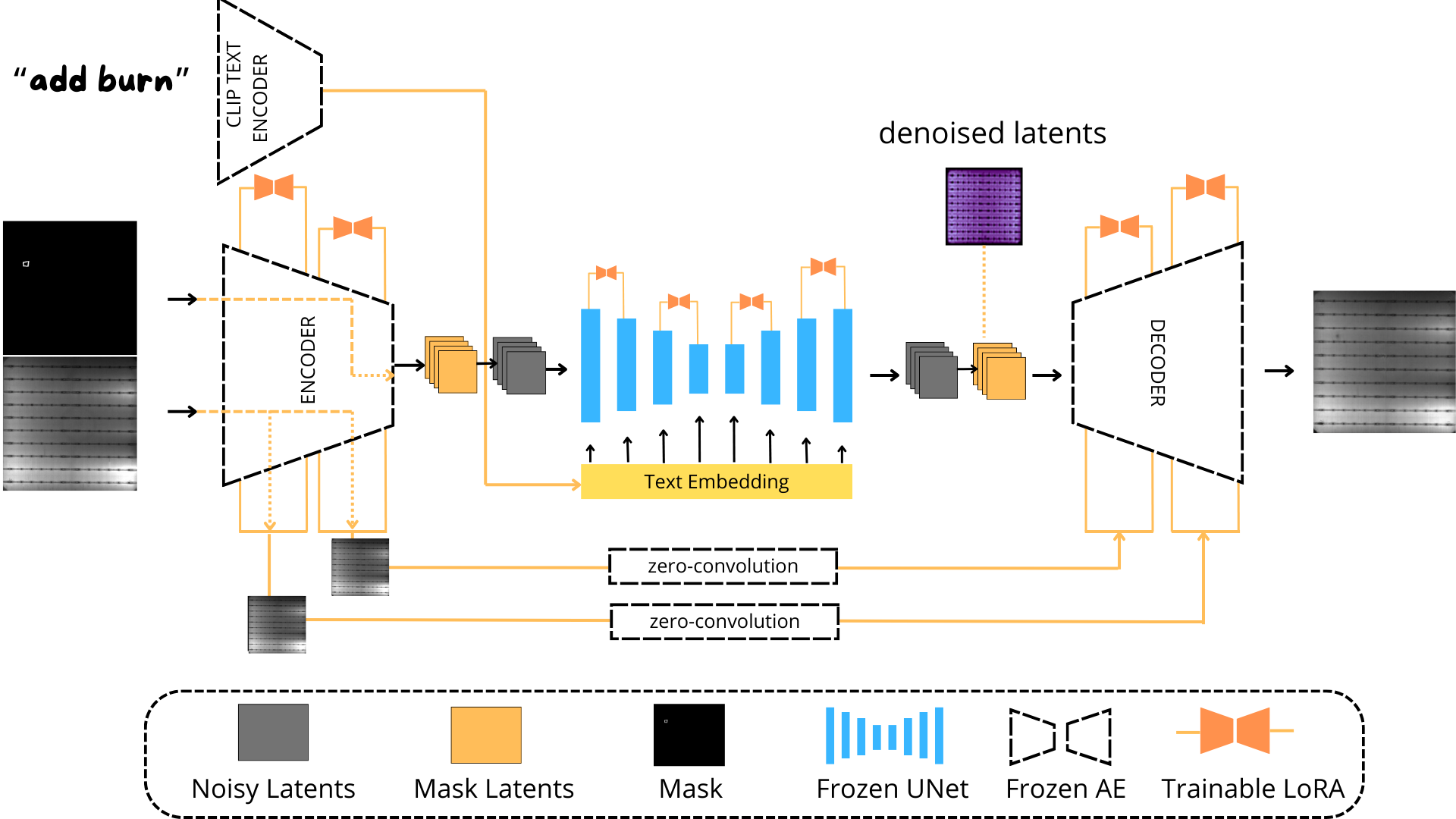}
\caption{\textbf{Ali-AUG architecture.} The architecture integrates three separate modules from the original latent diffusion models into a unified end-to-end network. Skip connections, Zero-Convs, and LoRA adapters are incorporated to retain input image details and ensure efficient mask-guided modification.}    \label{fig:ali_aug_architecture}
\end{figure}

This section introduces our novel approach for generating synthetic, labeled images using a single-step diffusion model, referred to as \textit{Ali-AUG} (see Figure \ref{fig:ali_aug_architecture}). Our model is based on  Parmar's model \cite{parmarOneStepImageTranslation2024} (see Figure \ref{fig:parmar_model}). However, while Parmar's approach focuses on generating an image from a sketch or canny image, our approach is designed to insert specific features into designated regions of an image given an input image. For example, it can add a defect to an object. This method utilizes the provided mask and is guided by a prompt to ensure the desired features are incorporated into specific areas of the input image without affecting the rest of the content, making it particularly useful for industrial applications.

\subsection{\textbf{Ali-AUG} for Text-to-Image diffusion.}  \leavevmode \\

We used Stable Diffusion as an example to demonstrate how we could edit images with a large pre-trained diffusion model. Stable Diffusion is composed of three main parts: an encoder, a U-Net \cite{ronneberger2015unetconvolutionalnetworksbiomedical}, and a decoder. The U-Net contains an encoder, a middle block, and a skip-connected decoder. Both the encoder and decoder consist of multiple blocks, with the encoder handling downsampling and the decoder performing upsampling. The model architecture also includes Vision Transformers (ViTs) that provide attention-based processing. In this framework, the text prompts (T) are encoded using the CLIP text encoder \cite{radford2021learningtransferablevisualmodels}, and the diffusion timesteps are encoded through a time encoder utilizing positional encoding \cite{zhang2023addingconditionalcontroltexttoimage}. The ViTs incorporate cross-attention mechanisms, allowing finer control of the prompt by modifying the internal attention maps of the diffusion model during inference alongside the self-attention mechanisms.

\textit{Ali-AUG} does not add any overhead to the current model; in fact, it only adds LoRA adapters in the encoder, decoder, and U-Net to make it efficient and possible to train even on graphics cards with 24 GB of VRAM. Additionally, skip connections have been added to preserve the input image information, which will be explained in detail below. Stable Diffusion uses a preprocessing method to convert 512 × 512 pixel-space images into smaller 64 × 64 latent images. In Ali-AUG, we first convert both the input image and the conditioning image (Mask) from 512 × 512 into a 64 × 64 feature space vector \cite{rombach2022highresolutionimagesynthesislatent} to match the size of Stable Diffusion using the same encoder.

\subsection{ Feature Extraction}
To achieve precise image synthesis, we utilize both the input image and the mask in our encoding process. The image and mask are encoded using a Stable Diffusion architecture enhanced with skip connections and LoRA modules. The encoding process involves several steps:

\textbf{Image and Mask Encoding}: Both the input image (I) and the mask (M) are encoded using the same encoder to extract relevant features. The encoder compresses the input image spatially while increasing the channel count ($512\times512\times3 \rightarrow 64\times64\times4$), enabling efficient processing. This step ensures that the essential characteristics of both the image and mask are captured, setting the stage for effective integration in subsequent stages.

\[ F_I = E(I), \quad F_M = E(M) \]

where \( E \) is the encoder that extracts features \( F_I \) from the input image and \( F_M \) from the mask.

\textbf{Skip Connections}: To preserve fine details from the input image, intermediate activations are extracted following each downsampling block within the encoder. These activations are added to the corresponding upsampling block of the decoder via a 1x1 zero convolution layer. This method ensures that fine details are retained throughout the image generation process.

\[ F = F_M + \text{ZeroConv}(F_I) \]

\textbf{LoRA Modules}: LoRA modules are employed to integrate mask-conditioned features into the targeted image area. These modules adapt the network weights to the new conditioning, reducing overfitting and fine-tuning time. This method also minimizes the number of parameters that need to be trained, enhancing the efficiency and transferability of the model. LoRA modules are added in the encoder, decoder, and U-Net to ensure effective integration of mask information (see Figure \ref{fig:ali_aug_architecture}).

\subsection{Preserving Input Details} \leavevmode \

A key challenge in multi-object and complex scene synthesis is the preservation of fine details. To address this, our approach involves compressing both the mask and the input images using the same encoder. This encoder spatially compresses the images and extracts relevant features, which may otherwise lead to a loss of intricate details.

To mitigate this loss, we employ skip connections between the encoder and decoder. Specifically, intermediate activations are extracted from each downsampling block within the encoder during the encoding of the input image. These activations are then processed via a 1x1 zero convolution layer \cite{zhangAddingConditionalControl2023} to ensure they retain essential details. These processed activations are subsequently fed into the corresponding upsampling blocks in the decoder. This method ensures that fine-grained details are preserved throughout the image generation process, significantly improving the quality and fidelity of the output \cite{parmarOneStepImageTranslation2024}.
\vspace{0.5cm}

\textbf{Dropping Input}: Dropout is applied to the input image (\(<30\%\)) during training, allowing the model to learn to generate images using only the mask and the prompt during inference. This enhances the model's versatility and applicability in various industrial settings by enabling it to generate defective images solely from the mask and the prompt.

\subsection{Training} \leavevmode \\

The training process for \textit{Ali-AUG} involves several key components to ensure high-quality image synthesis while maintaining efficiency. We utilized various hyperparameters, detailed in Table \ref{tab:hyperparameters}
, to fine-tune the model and ensure optimal performance.

\textbf{Basic Augmentation}: During training, we applied basic data augmentation techniques such as flipping, rotation, and color jittering to enhance the robustness of our model.

\textbf{Loss Functions}: To train our model, we utilize a combination of adversarial loss, reconstruction loss, and LPIPS (Learned Perceptual Image Patch Similarity) loss. The total loss \( \mathcal{L}_{total} \) is given by:

\[ \mathcal{L}_{total} = \lambda_{adv} \mathcal{L}_{adv} + \lambda_{rec} \mathcal{L}_{rec} + \lambda_{lpips} \mathcal{L}_{lpips} \]

The generator \( G \) aims to minimize this total loss:

\[
G^* = \arg \min_G \mathcal{L}_{total}
\]
\textbf{Adversarial Loss}: Ensures that the generated images are realistic. The adversarial loss is defined as:

\[ \mathcal{L}_{adv} = \mathbb{E}_{I, M, T} \left[ \log D(I) \right] + \mathbb{E}_{I, M, T} \left[ \log (1 - D(G(I, M, T))) \right] \]
where \( D \) is the discriminator network.

\textbf{Reconstruction Loss}: Ensures that the generated image \( I' \) closely matches the desired output based on the input image \( I \) and mask \( M \):

\[ \mathcal{L}_{rec} = \mathbb{E}_{I, M, T} \left[ \| I' - I_{target} \|_2 \right] \]
\textbf{LPIPS Loss}: Measures the perceptual similarity between the generated image and the target image:

\[ \mathcal{L}_{lpips} = \mathbb{E}_{I, M, T} \left[ \text{LPIPS}(I', I_{target}) \right] \]


With this setup, we can generate larger datasets using synthetic images, where the provided mask acts as the label, eliminating the need for manual re-labeling. By integrating masks as labels and leveraging advanced architectural elements like zero-convolution layers, our method efficiently generates high-quality synthetic images with precise feature placement. This efficiency allows us to quickly produce large amounts of labeled data, which is crucial for training compact models. For example, we can train lightweight object detection systems like YOLO (You Only Look Once) \cite{redmon2016you}, a state-of-the-art real-time object detection system. These compact models are well-suited for deployment on devices with limited computational capacity, extending the applicability of our approach to resource-constrained industrial environments.

\section{Experiments} \label{sec:experiments}

We conducted experiments to evaluate the effectiveness of our proposed method and compare it with existing models using the Frechet Inception Distance (FID) metric \cite{heusel2018ganstrainedtimescaleupdate}, following the clean-FID's implementation \cite{parmar2022aliasedresizingsurprisingsubtleties}. Additionally, we used two different data preparation strategies, Classification Accuracy Score (CAS) and Naive Augmentation Score (NAS), to further assess the models' performance as mentioned by Ravuri and Vinyals \cite{ravuri2019classificationaccuracyscoreconditional}. These strategies are explained in detail in Section \ref{sec:data_prep}.

We used multiple datasets for our experiments: a private industrial dataset of solar panel defects and Wood, Tile, Screw, and Leather from the MVTec AD \cite{8954181} dataset. These datasets are widely used for defect detection in industrial settings and include images of wood, tiles, screws, and leather surfaces, each with various defects closely resembling those used in real-world industrial applications.

\subsection{Metrics} \label{sec:metrics}
\subsubsection{Frechet Inception Distance (FID)}
The FID score measures the distance between the feature distributions of real and generated images, with lower scores indicating better performance.

\subsection{Data Preparation Strategies} \label{sec:data_prep}

\subsubsection{Classification Accuracy Score (CAS)}
For CAS, all training images are synthetic and generated using our method. The validation is performed on the validation dataset consisting of real images. This strategy helps assess the ability of models trained exclusively on synthetic data to generalize to real-world data.

\subsubsection{Naive Augmentation Score (NAS)}
For NAS, the training dataset is a mixture of synthetic and real images, and the validation is performed on real images. This strategy assesses the benefit of combining synthetic and real data during training to improve the model's performance on real-world data.

\subsection{Datasets}   \leavevmode \\

We conducted our experiments using both a private industrial dataset and public datasets to demonstrate the model's versatility and reproducibility.

\subsubsection{Private Industrial Dataset}

Initially, we conducted experiments on a private solar panel defects dataset, which cannot be publicly released. Due to the limited availability of labeled data---specifically, we only had \textbf{19 to 50 manually labeled images}---we leveraged our model to generate additional synthetic images to supplement the dataset. We created both paired and unpaired datasets for fine-tuning our model:

\begin{itemize}
    \item \textbf{Paired Dataset}: Consists of manually labeled images before and after modification with a condition and mask. This setup allows direct comparison and precise control over the modifications introduced by our model.
    \item \textbf{Unpaired Dataset}: Utilizes randomly selected defect-free images as inputs and the original defect images as outputs. Here, the model learns to introduce defects based on the provided masks and conditions without requiring exact image pairs.
\end{itemize}

By augmenting our limited labeled data with synthetic images generated by our model, we were able to create a more extensive dataset to validate the effectiveness and utility of the synthetic images in improving model performance. \textbf{This augmented dataset was crucial for assessing how well the synthetic images enhance the training of downstream models.} Samples of the industrial dataset used for these experiments are shown in Table~\ref{tab:industrial_samples}. Although we cannot share the solar panel dataset, we have replicated the same procedure using public datasets such as Wood, Tile, Screw, and Leather from the MVTec AD dataset \cite{8954181} to ensure reproducibility.

\begin{table}[H]
\centering
\begin{tabular}{cccc}
\hline
Input Image & Source & Model Output & Target \\ 
\includegraphics[width=0.2\textwidth]{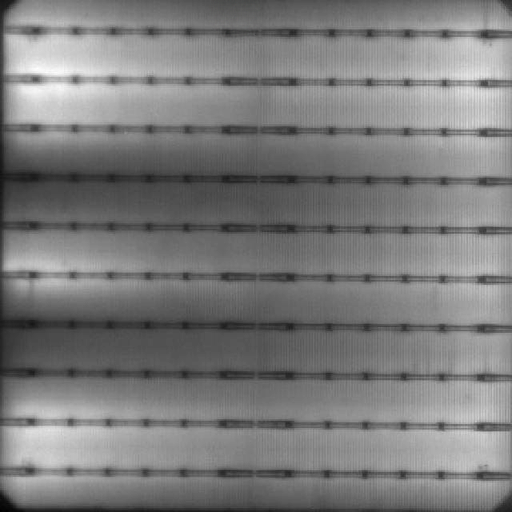} & \includegraphics[width=0.2\textwidth]{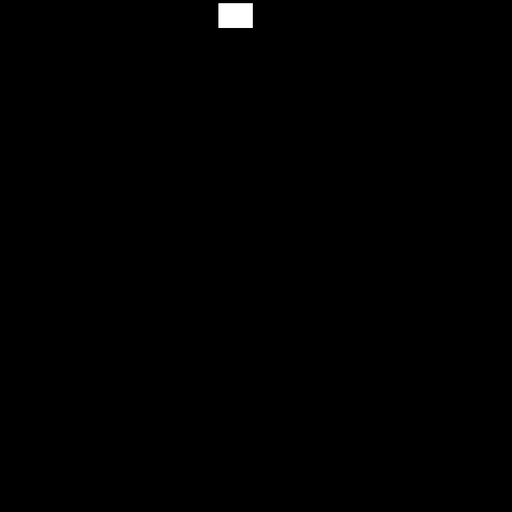} & \includegraphics[width=0.2\textwidth]{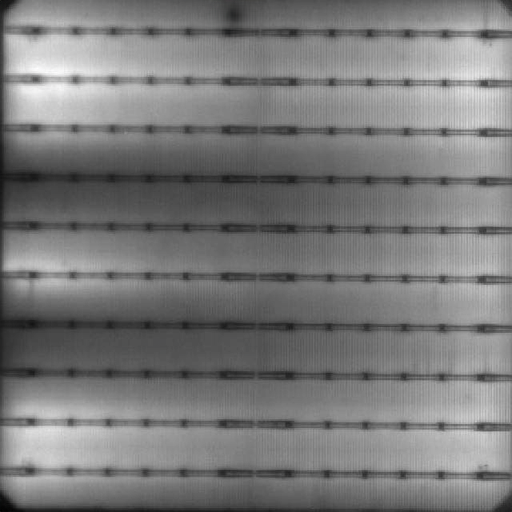} & \includegraphics[width=0.2\textwidth]{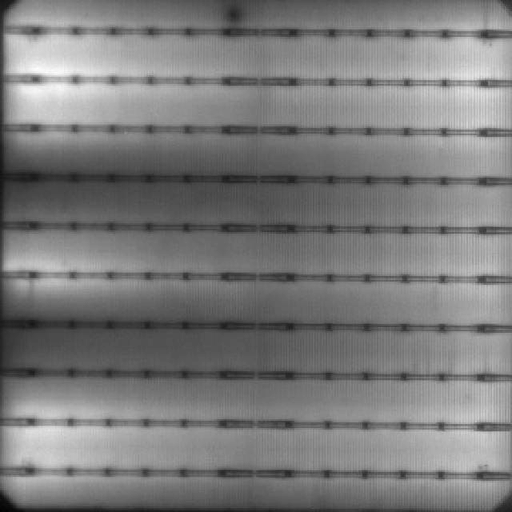} \\ 
\includegraphics[width=0.2\textwidth]{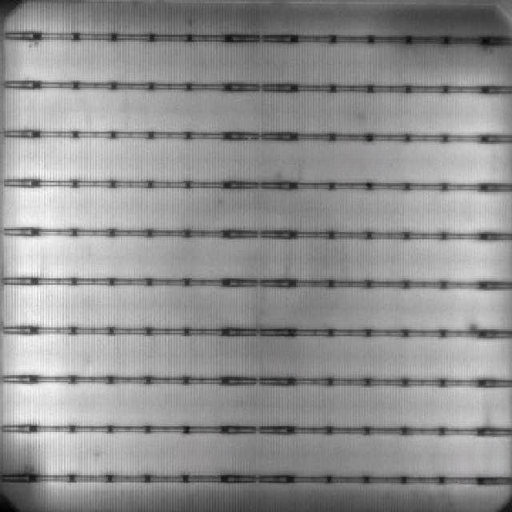} & \includegraphics[width=0.2\textwidth]{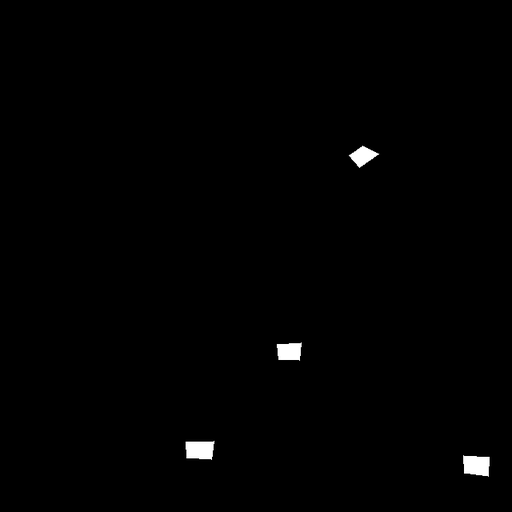} & \includegraphics[width=0.2\textwidth]{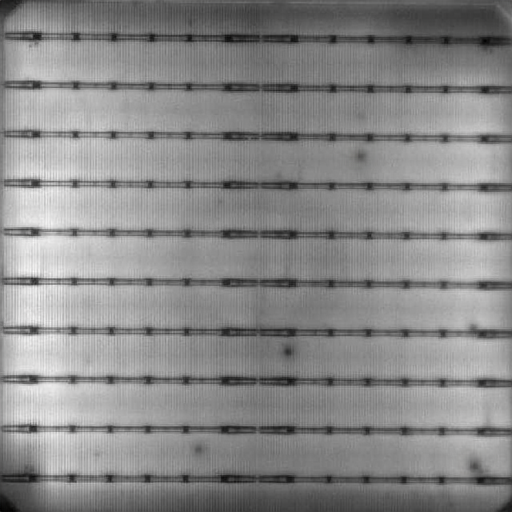} & \includegraphics[width=0.2\textwidth]{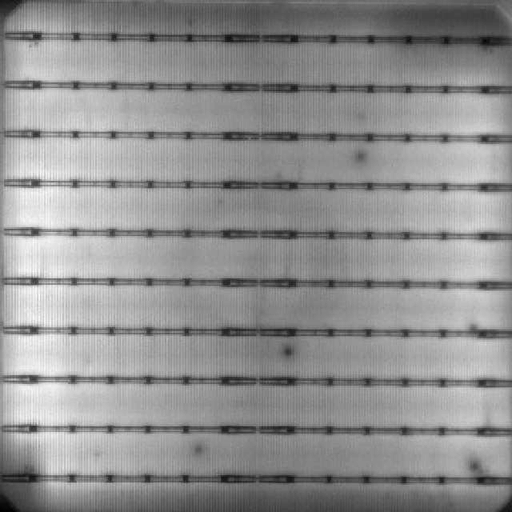} \\ \hline
\end{tabular}
\caption{Images generated by the industrial paired dataset after 30 minutes of training with the prompt "add burn".}

\label{tab:industrial_samples}
\end{table}

\subsubsection{Private Industrial Paired Training}

In our private industrial dataset, we have a total of 900 images, of which only 32 are defective. Using traditional models such as YOLO and autoencoders with conventional data augmentation techniques, the models often resulted in a high number of false positives. To address this, we utilized Ali-AUG to generate the CAS and NAS datasets to validate our model's robustness. This approach achieved a FID score of 5.5, as shown in Table \ref{tab:fid_results}. Subsequently, we used a YOLO model on these datasets for defect detection, demonstrating improved results compared to traditional data augmentation methods. The detailed results are presented in Table \ref{tab:training_results_panels}.

\begin{table}[H]
    \centering
    \renewcommand{\arraystretch}{1.1}
    \setlength{\tabcolsep}{5pt}
    \begin{small}
    \begin{tabular}{@{}>{\centering\arraybackslash}m{0.26\textwidth} >{\centering\arraybackslash}m{0.16\textwidth} >{\centering\arraybackslash}m{0.16\textwidth} >{\centering\arraybackslash}m{0.16\textwidth} >{\centering\arraybackslash}m{0.16\textwidth}@{}}
        \toprule
        \textbf{Name} & \textbf{D\_S} & \textbf{D\_S\_AUG} & \textbf{CAS} & \textbf{NAS} \\ \midrule
        metrics/precision     & 0.551 & 0.722 & 0.821 & \textbf{0.908} \\ \hdashline
        metrics/recall        & 0.523 & 0.715 & 0.825 & \textbf{0.875} \\ \hdashline
        metrics/mAP50         & 0.512 & 0.712 & 0.885 & \textbf{0.933} \\ \hdashline
        metrics/mAP50\_95     & 0.2144 & 0.265 & 0.323 & \textbf{0.411} \\ \bottomrule
    \end{tabular}
    \end{small}
    \caption{Training metrics results for different experimental setups on the \textbf{Industrial panels} paired dataset. 
    \textbf{D\_S}: A simple split of the dataset into 70\% training and 30\% testing. 
    \textbf{D\_S\_AUG}: Similar to Data\_Split but with simple data augmentation applied. 
    \textbf{CAS} and \textbf{NAS}: Only images generated by our model were used.
    All models were validated on the same 30\% split of the data. Augmentation was performed only on the 70\% training data. A total of 19 images were used.}
    \label{tab:training_results_panels}
\end{table}

\subsubsection{FID Results}

\begin{table}[H]
    \centering
    \renewcommand{\arraystretch}{1.1}
    \setlength{\tabcolsep}{5pt}
    \begin{small}
    \begin{tabular}{@{}>{\centering\arraybackslash}m{0.26\textwidth} >{\centering\arraybackslash}m{0.16\textwidth} >{\centering\arraybackslash}m{0.16\textwidth} >{\centering\arraybackslash}m{0.16\textwidth}@{}}
        \toprule
        \textbf{Dataset} & \textbf{Ours} & \textbf{Single Step \cite{parmarOneStepImageTranslation2024}} & \textbf{BrushNet \cite{juBrushNetPlugandPlayImage2024}} \\ \midrule
        Wood & \textbf{65.3} & 80.9 & 67.1 \\ \hdashline
        Tile & \textbf{95.2} & 110.1 & 101.2 \\ \hdashline
        Screw & \textbf{35.4} & 46.2 & 38.8 \\ \hdashline
        Leather & 39.1 & 42.1 & \textbf{38.2} \\ \hdashline
        Industrial panels & \textbf{5.5} & 12.1 & 6.1 \\  \bottomrule
    \end{tabular}
    \end{small}
    \caption{FID scores for the Industrial panels, Wood, Tile, Screw, and Leather datasets across different models (Ours, Single Step, BrushNet).}
    \label{tab:fid_results}
\end{table}

\subsubsection{Public Dataset}

To enable other researchers to validate our approach and methodology, we conducted extensive experiments on public datasets, including Wood, Tile, Screw, and Leather from the MVTec AD dataset \cite{8954181}. While our implementation is proprietary, we provide comprehensive methodological details and experimental results that should allow researchers to understand and build upon our approach. We followed a similar process of generating synthetic images and creating paired and unpaired datasets for these experiments. The results from these public datasets are presented in Table \ref{tab:fid_results}, demonstrating the model's performance and validating our methodology through standard benchmarks and metrics.

Through these experiments, we highlight the model's capability to generate high-quality synthetic images, perform efficient data augmentation, and maintain fast training times. The validation results using CAS and NAS metrics further confirm the effectiveness of our approach, providing a robust framework for data augmentation in various industrial applications.

\subsubsection{Public Dataset Paired Training}
For paired datasets, the model is fine-tuned to meticulously follow the pattern and details of the original image, embedding the desired feature into the marked area. This setting ensures high fidelity and precise feature insertion. When it is not possible to obtain the dataset with an image before and after modification given a condition and mask, we have used an inpainting model \cite{suvorov2021resolutionrobustlargemaskinpainting} to remove defects from the image. On paired datasets, the model learns really quickly. For example, with 15 images of screws, after 1000 steps with a batch size of 1, the model is already capable of generating defects correctly (see Table \ref{tab:screw_table}).

\begin{table}[H]
\centering
\begin{tabular}{cccc}
\hline
   Input Image & Mask & Model Output & Ground truth \\
\includegraphics[width=0.2\textwidth]{imgs/tornillos_generated/178900/media_images_train_input.png} & \includegraphics[width=0.2\textwidth]{imgs/tornillos_generated/178900/media_images_train_source.png} & \includegraphics[width=0.2\textwidth]{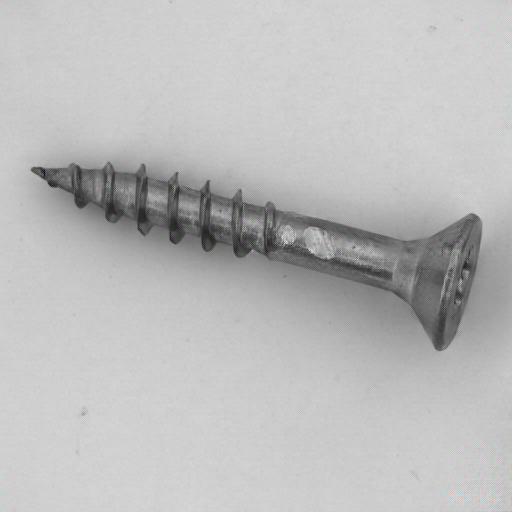} & \includegraphics[width=0.2\textwidth]{imgs/tornillos_generated/178900/media_images_train_target.png} \\

\hline
\end{tabular}
\caption{Comparison of model output against ground truth for screw images. The images show the mask used, the input image, the model's generated output, and the actual ground truth image. This demonstrates the model's ability to generate defects accurately after 1,000 training steps with a small paired dataset of 15 images with the prompt "add scratch."}
\label{tab:screw_table}
\end{table}

\subsubsection{Public Dataset Unpaired Training}

In scenarios where paired data is unavailable and challenging to create, our model leverages unpaired training techniques. Since we have defect-labeled images, we use these labels as masks and the defective image as the target. For the input, a defect-free image is selected randomly. This approach ensures that the model continues to train quickly while maintaining high-quality results (see dataset example in Table \ref{tab:unpaired_label}). Additional synthetically generated images can be found in Table \ref{tab:training_unpaired_res_2}.

Once the dataset was generated and the model trained, we created the CAS and NAS datasets. We then followed the same procedure as with the private solar panel defects dataset, training YOLOv8-seg on these datasets for segmentation instead of object detection. This approach aimed to demonstrate improvements across different scenarios. The results can be seen in Tables \ref{tab:training_results_binary} and \ref{tab:training_results_multiclass}.

\begin{table}[H]
    \centering
    \renewcommand{\arraystretch}{1.1}
    \setlength{\tabcolsep}{5pt}
    \begin{small}
    \begin{tabular}{@{}>{\centering\arraybackslash}m{0.26\textwidth} >{\centering\arraybackslash}m{0.16\textwidth} >{\centering\arraybackslash}m{0.16\textwidth} >{\centering\arraybackslash}m{0.16\textwidth} >{\centering\arraybackslash}m{0.16\textwidth}@{}}
        \toprule
        \textbf{Name} & \textbf{D\_S} & \textbf{D\_S\_AUG} & \textbf{CAS} & \textbf{NAS} \\ \midrule
        metrics/precision(B)   & 0.479 & 0.549 & \textbf{0.971} & 0.970 \\ \hdashline
        metrics/recall(B)      & 0.655 & 0.715 & 0.936 & \textbf{0.998} \\ \hdashline
        metrics/mAP50(B)       & 0.627 & 0.712 & \textbf{0.995} & \textbf{0.995} \\ \hdashline
        metrics/mAP50\_95(B)    & 0.211 & 0.265 & 0.748 & \textbf{0.803} \\ \bottomrule
    \end{tabular}
    \end{small}
    \caption{Training results for binary classification metrics on the \textbf{Tile} dataset. 
    \textbf{D\_S}: A simple split of the dataset into 70\% training and 30\% testing. 
    \textbf{D\_S\_AUG}: Similar to Data\_Split but with simple data augmentation applied. 
    \textbf{CAS} and \textbf{NAS}: Only images generated by our model were used. 
    All models were validated on the same 30\% split of the data. Augmentation was performed only on the 70\% training data. A total of 19 images were used.}
    \label{tab:training_results_binary}
\end{table}

\begin{table}[H]
    \centering
    \renewcommand{\arraystretch}{1.1}
    \setlength{\tabcolsep}{5pt}
    \begin{small}
    \begin{tabular}{@{}>{\centering\arraybackslash}m{0.26\textwidth} >{\centering\arraybackslash}m{0.16\textwidth} >{\centering\arraybackslash}m{0.16\textwidth} >{\centering\arraybackslash}m{0.16\textwidth} >{\centering\arraybackslash}m{0.16\textwidth}@{}}
        \toprule
        \textbf{Name} & \textbf{D\_S} & \textbf{D\_S\_AUG} & \textbf{CAS} & \textbf{NAS} \\ \midrule
        metrics/precision(M)   & 0.517 & 0.586 & \textbf{0.971} & 0.970 \\ \hdashline
        metrics/recall(M)      & 0.722 & 0.819 & 0.936 & \textbf{0.998} \\ \hdashline
        metrics/mAP50(M)       & 0.660 & 0.761 & \textbf{0.995} & \textbf{0.995} \\ \hdashline
        metrics/mAP50\_95(M)    & 0.447 & 0.529 & 0.761 & \textbf{0.818} \\ \bottomrule
    \end{tabular}
    \end{small}
    \caption{Training results for multiclass classification metrics on the \textbf{Tile} dataset. 
    \textbf{D\_S}: A simple split of the dataset into 70\% training and 30\% testing. 
    \textbf{D\_S\_AUG}: Similar to Data\_Split but with simple data augmentation applied. 
    \textbf{CAS} and \textbf{NAS}: Only images generated by our model were used. 
    All models were validated on the same 30\% split of the data. Augmentation was performed only on the 70\% training data. A total of 19 images were used.}
    \label{tab:training_results_multiclass}
\end{table}

\section{Discussion}  \label{sec:discussion}

\subsection{Conclusions}  \leavevmode \\

This study introduces Ali-AUG, an innovative single-step diffusion model that significantly enhances data augmentation capabilities creating labeled images, particularly in industrial applications where labeled data is scarce or expensive. The model demonstrates remarkable efficiency, enabling rapid learning with minimal examples and reducing computational demands through its use of LoRA modules and skip connections. Its versatility across both paired and unpaired training settings, coupled with its ability to generate high-quality images with precise feature insertion, provides a powerful tool for improving the performance of deep learning models in various industrial scenarios. Experimental validation across multiple datasets confirms the model's superiority in generating realistic, defect-enhanced images while maintaining rapid single-step inference. By offering a cost-effective solution to the chronic shortage of labeled defect data and potentially supporting continuous learning paradigms, Ali-AUG represents a significant advancement in the field of synthetic data generation for industrial AI applications.

\subsection{Limitation and Future Work}  \leavevmode \\

Despite the advantages of our proposed method, several limitations exist. As with any generative model, the images produced by our approach can sometimes be incorrect or invalid. There are instances where the generated images might not accurately reflect the intended defect or characteristic, leading to potential inaccuracies in the labels. These issues arise from the inherent challenges of conditional image generation and the complexity of faithfully reproducing specific features in varied contexts.

Our model is designed to aid in accelerating data augmentation and labeling processes, but it is not a replacement for meticulous manual verification. Users should employ this tool to supplement their existing workflows, providing a means to rapidly generate large volumes of synthetic data, which can then be reviewed and corrected as needed. This approach is precious in scenarios where labeled data is scarce or expensive, substantially improving efficiency.

For future work, we propose the following directions to enhance the robustness and utility of our method:

\textbf{Improved Validation Mechanisms:} Developing automated validation techniques to assess the quality and accuracy of generated images could significantly reduce the need for manual review. Integrating machine learning models to detect and flag potential inaccuracies in real time would enhance the reliability of the generated dataset.

\textbf{Scalability and Adaptability:} Enhancing the model's capability to handle even smaller and more diverse datasets would be beneficial. Researching ways to generalize the model across different domains without extensive retraining can make it more adaptable to various industrial applications, enabling effective performance even with few-shot learning.



\section*{Acknowledgements}
This work has been partially funded by the IKUN project with reference code KK-2024/00064, titled "Large Multimodal Models for Quality Assurance and Operator Support in Intelligent Industry," for the period 2024-2025; by grant PID2021-123988OB-C31 funded by MICIU/AEI /10.13039/501100011033/; and by `ERDF A way of making Europe'.

\bibliography{references}

\begin{thebibliography}{10}

\bibitem{kumar2021binary}
Teerath Kumar, Jinbae Park, Muhammad~Salman Ali, AFM~Shahab Uddin, Jong~Hwan Ko, and Sung-Ho Bae.
\newblock Binary-classifiers-enabled filters for semi-supervised learning.
\newblock {\em IEEE Access}, 9:167663--167673, 2021.

\bibitem{kumar2023imagedataaugmentationapproaches}
Teerath Kumar, Alessandra Mileo, Rob Brennan, and Malika Bendechache.
\newblock Image data augmentation approaches: A comprehensive survey and future directions, 2023.

\bibitem{yun2019cutmixregularizationstrategytrain}
Sangdoo Yun, Dongyoon Han, Seong~Joon Oh, Sanghyuk Chun, Junsuk Choe, and Youngjoon Yoo.
\newblock Cutmix: Regularization strategy to train strong classifiers with localizable features, 2019.

\bibitem{liu2021pdganprobabilisticdiversegan}
Hongyu Liu, Ziyu Wan, Wei Huang, Yibing Song, Xintong Han, and Jing Liao.
\newblock Pd-gan: Probabilistic diverse gan for image inpainting, 2021.

\bibitem{zheng2022cmganimageinpaintingcascaded}
Haitian Zheng, Zhe Lin, Jingwan Lu, Scott Cohen, Eli Shechtman, Connelly Barnes, Jianming Zhang, Ning Xu, Sohrab Amirghodsi, and Jiebo Luo.
\newblock Cm-gan: Image inpainting with cascaded modulation gan and object-aware training, 2022.

\bibitem{peng2021generating}
Jialun Peng, Dong Liu, Songcen Xu, and Houqiang Li.
\newblock Generating diverse structure for image inpainting with hierarchical vq-vae.
\newblock In {\em Proceedings of the IEEE/CVF conference on computer vision and pattern recognition}, pages 10775--10784, 2021.

\bibitem{zheng2019pluralisticimagecompletion}
Chuanxia Zheng, Tat-Jen Cham, and Jianfei Cai.
\newblock Pluralistic image completion, 2019.

\bibitem{Shao2019GenerativeAN}
Siyu Shao, Pu~Wang, and Ruqiang Yan.
\newblock Generative adversarial networks for data augmentation in machine fault diagnosis.
\newblock {\em Comput. Ind.}, 106:85--93, 2019.

\bibitem{huang2024diffusionmodelbasedimageediting}
Yi~Huang, Jiancheng Huang, Yifan Liu, Mingfu Yan, Jiaxi Lv, Jianzhuang Liu, Wei Xiong, He~Zhang, Shifeng Chen, and Liangliang Cao.
\newblock Diffusion model-based image editing: A survey, 2024.

\bibitem{zhangAddingConditionalControl2023}
Lvmin Zhang, Anyi Rao, and Maneesh Agrawala.
\newblock Adding {{Conditional Control}} to {{Text-to-Image Diffusion Models}}.

\bibitem{juBrushNetPlugandPlayImage2024}
Xuan Ju, Xian Liu, Xintao Wang, Yuxuan Bian, Ying Shan, and Qiang Xu.
\newblock {{BrushNet}}: {{A Plug-and-Play Image Inpainting Model}} with {{Decomposed Dual-Branch Diffusion}}.

\bibitem{Wu_2023_ICCV}
Chen~Henry Wu and Fernando De~la Torre.
\newblock A latent space of stochastic diffusion models for zero-shot image editing and guidance.
\newblock In {\em Proceedings of the IEEE/CVF International Conference on Computer Vision (ICCV)}, pages 7378--7387, October 2023.

\bibitem{wassermanPaintInpaintLearning2024}
Navve Wasserman, Noam Rotstein, Roy Ganz, and Ron Kimmel.
\newblock Paint by {{Inpaint}}: {{Learning}} to {{Add Image Objects}} by {{Removing Them First}}.

\bibitem{lugmayrRePaintInpaintingUsing2022}
Andreas Lugmayr, Martin Danelljan, Andres Romero, Fisher Yu, Radu Timofte, and Luc Van~Gool.
\newblock {{RePaint}}: {{Inpainting}} using {{Denoising Diffusion Probabilistic Models}}.

\bibitem{corneanuLatentPaintImageInpainting2024}
Ciprian Corneanu, Raghudeep Gadde, and Aleix~M Martinez.
\newblock {{LatentPaint}}: {{Image Inpainting}} in {{Latent Space}} with {{Diffusion Models}}.
\newblock In {\em 2024 {{IEEE}}/{{CVF Winter Conference}} on {{Applications}} of {{Computer Vision}} ({{WACV}})}, pages 4322--4331. IEEE.

\bibitem{zhang2023addingconditionalcontroltexttoimage}
Lvmin Zhang, Anyi Rao, and Maneesh Agrawala.
\newblock Adding conditional control to text-to-image diffusion models, 2023.

\bibitem{parmarOneStepImageTranslation2024}
Gaurav Parmar, Taesung Park, Srinivasa Narasimhan, and Jun-Yan Zhu.
\newblock One-{{Step Image Translation}} with {{Text-to-Image Models}}.

\bibitem{hu2021loralowrankadaptationlarge}
Edward~J. Hu, Yelong Shen, Phillip Wallis, Zeyuan Allen-Zhu, Yuanzhi Li, Shean Wang, Lu~Wang, and Weizhu Chen.
\newblock Lora: Low-rank adaptation of large language models, 2021.

\bibitem{xie2023smartbrush}
Shaoan Xie, Zhifei Zhang, Zhe Lin, Tobias Hinz, and Kun Zhang.
\newblock Smartbrush: Text and shape guided object inpainting with diffusion model.
\newblock In {\em Proceedings of the IEEE/CVF Conference on Computer Vision and Pattern Recognition}, pages 22428--22437, 2023.

\bibitem{xie2023dreaminpaintertextguidedsubjectdrivenimage}
Shaoan Xie, Yang Zhao, Zhisheng Xiao, Kelvin C.~K. Chan, Yandong Li, Yanwu Xu, Kun Zhang, and Tingbo Hou.
\newblock Dreaminpainter: Text-guided subject-driven image inpainting with diffusion models, 2023.

\bibitem{yang2022image}
Suorong Yang, Weikang Xiao, Mengchen Zhang, Suhan Guo, Jian Zhao, and Furao Shen.
\newblock Image data augmentation for deep learning: A survey.
\newblock {\em arXiv preprint arXiv:2204.08610}, 2022.

\bibitem{ren2019mask}
Yinhao Ren, Zhe Zhu, Yingzhou Li, and Joseph Lo.
\newblock Mask embedding in conditional gan for guided synthesis of high resolution images.
\newblock {\em arXiv preprint arXiv:1907.01710}, 2019.

\bibitem{ho2020denoisingdiffusionprobabilisticmodels}
Jonathan Ho, Ajay Jain, and Pieter Abbeel.
\newblock Denoising diffusion probabilistic models, 2020.

\bibitem{song2022denoisingdiffusionimplicitmodels}
Jiaming Song, Chenlin Meng, and Stefano Ermon.
\newblock Denoising diffusion implicit models, 2022.

\bibitem{openai_dalle3}
OpenAI.
\newblock Dall·e 3.
\newblock \url{https://openai.com/index/dall-e-3/}, 2024.
\newblock Accessed: 2024-05-24.

\bibitem{midjourney_home}
Midjourney.
\newblock Midjourney.
\newblock \url{https://www.midjourney.com/home}, 2024.
\newblock Accessed: 2024-05-24.

\bibitem{rombachHighResolutionImageSynthesis2022}
Robin Rombach, Andreas Blattmann, Dominik Lorenz, Patrick Esser, and Björn Ommer.
\newblock High-{{Resolution Image Synthesis}} with {{Latent Diffusion Models}}.

\bibitem{islamDIFFUSEMIXLabelPreservingData}
Khawar Islam, Muhammad~Zaigham Zaheer, Arif Mahmood, and Karthik Nandakumar.
\newblock {{DIFFUSEMIX}}: {{Label-Preserving Data Augmentation}} with {{Diffusion Models}}.

\bibitem{trabuccoEffectiveDataAugmentation2023}
Brandon Trabucco, Kyle Doherty, Max Gurinas, and Ruslan Salakhutdinov.
\newblock Effective {{Data Augmentation With Diffusion Models}}.

\bibitem{reed2016generativeadversarialtextimage}
Scott Reed, Zeynep Akata, Xinchen Yan, Lajanugen Logeswaran, Bernt Schiele, and Honglak Lee.
\newblock Generative adversarial text to image synthesis, 2016.

\bibitem{kim2017learning}
Taeksoo Kim, Moonsu Cha, Hyunsoo Kim, Jung~Kwon Lee, and Jiwon Kim.
\newblock Learning to discover cross-domain relations with generative adversarial networks, 2017.

\bibitem{taigman2016unsupervised}
Yaniv Taigman, Adam Polyak, and Lior Wolf.
\newblock Unsupervised cross-domain image generation, 2016.

\bibitem{mouT2IAdapterLearningAdapters2023}
Chong Mou, Xintao Wang, Liangbin Xie, Yanze Wu, Jian Zhang, Zhongang Qi, Ying Shan, and Xiaohu Qie.
\newblock {{T2I-Adapter}}: {{Learning Adapters}} to {{Dig}} out {{More Controllable Ability}} for {{Text-to-Image Diffusion Models}}.

\bibitem{avrahami2022blended}
Omri Avrahami, Dani Lischinski, and Ohad Fried.
\newblock Blended diffusion for text-driven editing of natural images.
\newblock In {\em Proceedings of the IEEE/CVF conference on computer vision and pattern recognition}, pages 18208--18218, 2022.

\bibitem{rombach2022highresolutionimagesynthesislatent}
Robin Rombach, Andreas Blattmann, Dominik Lorenz, Patrick Esser, and Björn Ommer.
\newblock High-resolution image synthesis with latent diffusion models, 2022.

\bibitem{manukyan2024hdpainterhighresolutionpromptfaithfultextguided}
Hayk Manukyan, Andranik Sargsyan, Barsegh Atanyan, Zhangyang Wang, Shant Navasardyan, and Humphrey Shi.
\newblock Hd-painter: High-resolution and prompt-faithful text-guided image inpainting with diffusion models, 2024.

\bibitem{chen2024virtualmodelgeneratingobjectidretentivehumanobject}
Binghui Chen, Chongyang Zhong, Wangmeng Xiang, Yifeng Geng, and Xuansong Xie.
\newblock Virtualmodel: Generating object-id-retentive human-object interaction image by diffusion model for e-commerce marketing, 2024.

\bibitem{wang2023imagen}
Su~Wang, Chitwan Saharia, Ceslee Montgomery, Jordi Pont-Tuset, Shai Noy, Stefano Pellegrini, Yasumasa Onoe, Sarah Laszlo, David~J Fleet, Radu Soricut, et~al.
\newblock Imagen editor and editbench: Advancing and evaluating text-guided image inpainting.
\newblock In {\em Proceedings of the IEEE/CVF conference on computer vision and pattern recognition}, pages 18359--18369, 2023.

\bibitem{brooksInstructPix2PixLearningFollow2023}
Tim Brooks, Aleksander Holynski, and Alexei~A. Efros.
\newblock {{InstructPix2Pix}}: {{Learning}} to {{Follow Image Editing Instructions}}.
\newblock In {\em 2023 {{IEEE}}/{{CVF Conference}} on {{Computer Vision}} and {{Pattern Recognition}} ({{CVPR}})}, pages 18392--18402. IEEE.

\bibitem{fuGuidingInstructionbasedImage2024}
Tsu-Jui Fu, Wenze Hu, Xianzhi Du, William~Yang Wang, Yinfei Yang, and Zhe Gan.
\newblock Guiding {{Instruction-based Image Editing}} via {{Multimodal Large Language Models}}.

\bibitem{lu2022dpm}
Cheng Lu, Yuhao Zhou, Fan Bao, Jianfei Chen, Chongxuan Li, and Jun Zhu.
\newblock Dpm-solver: A fast ode solver for diffusion probabilistic model sampling in around 10 steps.
\newblock {\em Advances in Neural Information Processing Systems}, 35:5775--5787, 2022.

\bibitem{meng2023distillation}
Chenlin Meng, Robin Rombach, Ruiqi Gao, Diederik~P. Kingma, Stefano Ermon, Jonathan Ho, and Tim Salimans.
\newblock On distillation of guided diffusion models, 2023.

\bibitem{salimans2022progressive}
Tim Salimans and Jonathan Ho.
\newblock Progressive distillation for fast sampling of diffusion models, 2022.

\bibitem{sauer2023adversarial}
Axel Sauer, Dominik Lorenz, Andreas Blattmann, and Robin Rombach.
\newblock Adversarial diffusion distillation, 2023.

\bibitem{wang2023prolificdreamer}
Zhengyi Wang, Cheng Lu, Yikai Wang, Fan Bao, Chongxuan Li, Hang Su, and Jun Zhu.
\newblock Prolificdreamer: High-fidelity and diverse text-to-3d generation with variational score distillation, 2023.

\bibitem{ronneberger2015unetconvolutionalnetworksbiomedical}
Olaf Ronneberger, Philipp Fischer, and Thomas Brox.
\newblock U-net: Convolutional networks for biomedical image segmentation, 2015.

\bibitem{radford2021learningtransferablevisualmodels}
Alec Radford, Jong~Wook Kim, Chris Hallacy, Aditya Ramesh, Gabriel Goh, Sandhini Agarwal, Girish Sastry, Amanda Askell, Pamela Mishkin, Jack Clark, Gretchen Krueger, and Ilya Sutskever.
\newblock Learning transferable visual models from natural language supervision, 2021.

\bibitem{redmon2016you}
Joseph Redmon, Santosh Divvala, Ross Girshick, and Ali Farhadi.
\newblock You only look once: Unified, real-time object detection.
\newblock In {\em Proceedings of the IEEE conference on computer vision and pattern recognition}, pages 779--788, 2016.

\bibitem{heusel2018ganstrainedtimescaleupdate}
Martin Heusel, Hubert Ramsauer, Thomas Unterthiner, Bernhard Nessler, and Sepp Hochreiter.
\newblock Gans trained by a two time-scale update rule converge to a local nash equilibrium, 2018.

\bibitem{parmar2022aliasedresizingsurprisingsubtleties}
Gaurav Parmar, Richard Zhang, and Jun-Yan Zhu.
\newblock On aliased resizing and surprising subtleties in gan evaluation, 2022.

\bibitem{ravuri2019classificationaccuracyscoreconditional}
Suman Ravuri and Oriol Vinyals.
\newblock Classification accuracy score for conditional generative models, 2019.

\bibitem{8954181}
Paul Bergmann, Michael Fauser, David Sattlegger, and Carsten Steger.
\newblock Mvtec ad — a comprehensive real-world dataset for unsupervised anomaly detection.
\newblock In {\em 2019 IEEE/CVF Conference on Computer Vision and Pattern Recognition (CVPR)}, pages 9584--9592, 2019.

\bibitem{suvorov2021resolutionrobustlargemaskinpainting}
Roman Suvorov, Elizaveta Logacheva, Anton Mashikhin, Anastasia Remizova, Arsenii Ashukha, Aleksei Silvestrov, Naejin Kong, Harshith Goka, Kiwoong Park, and Victor Lempitsky.
\newblock Resolution-robust large mask inpainting with fourier convolutions, 2021.

\end{thebibliography}
\bibliographystyle{unsrt}



\newpage
\appendix
\section{Additional Results}
\label{sec:additional_results}

\begin{table}[H]
\centering
\begin{tabular}{cccc}
\hline
Input Image  & Mask & Model Output & Ground truth \\ 
\includegraphics[width=0.2\textwidth]{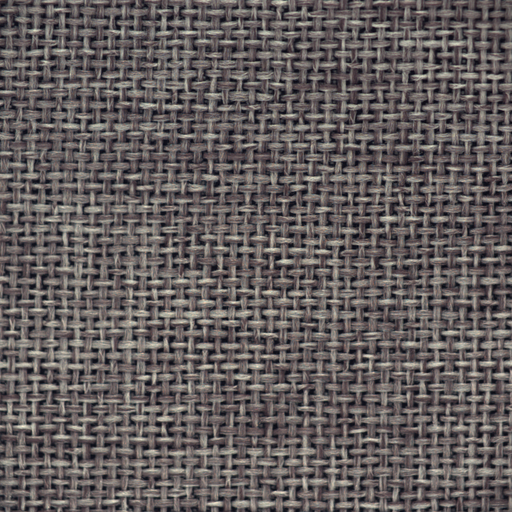} & \includegraphics[width=0.2\textwidth]{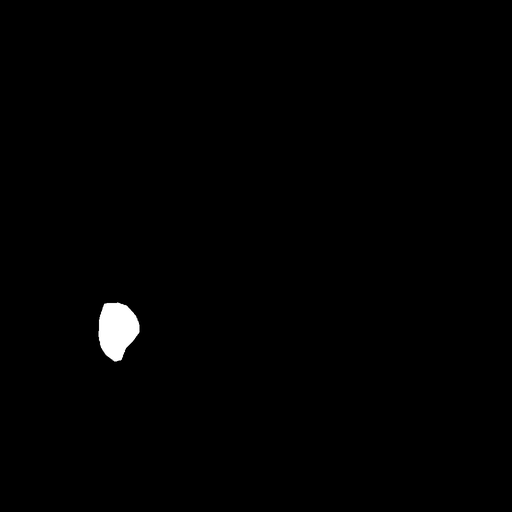} & \includegraphics[width=0.2\textwidth]{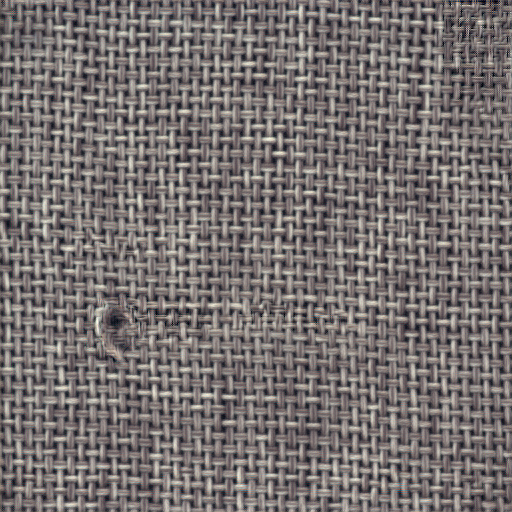} & \includegraphics[width=0.2\textwidth]{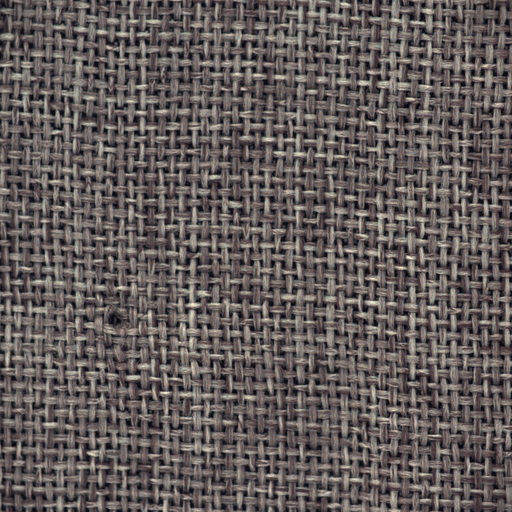} \\
\includegraphics[width=0.2\textwidth]{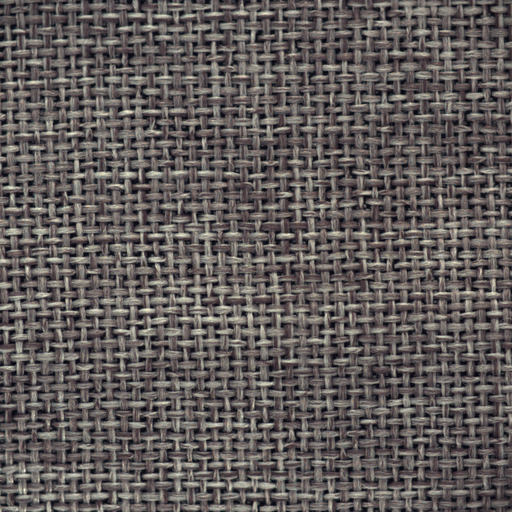} & \includegraphics[width=0.2\textwidth]{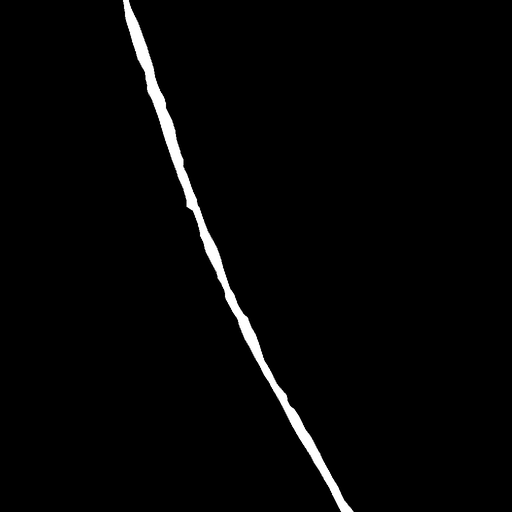} & \includegraphics[width=0.2\textwidth]{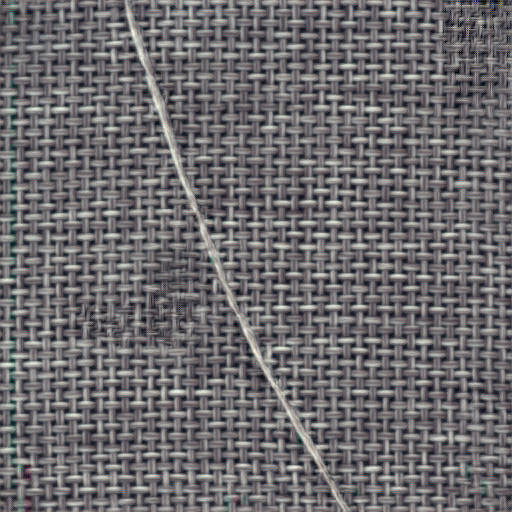} & \includegraphics[width=0.2\textwidth]{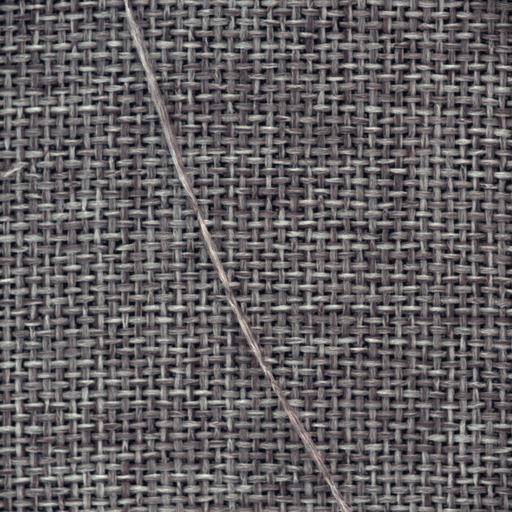} \\

\includegraphics[width=0.2\textwidth]{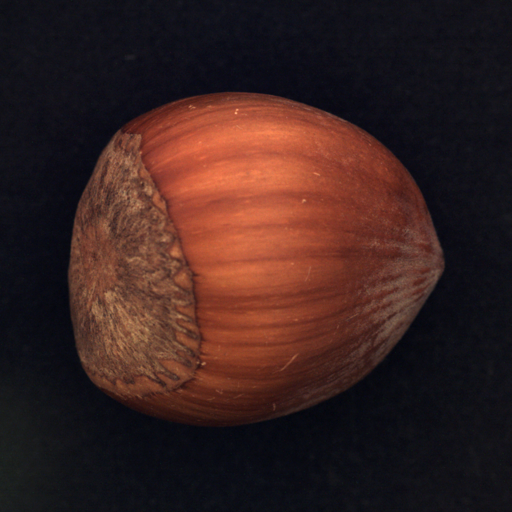} & \includegraphics[width=0.2\textwidth]{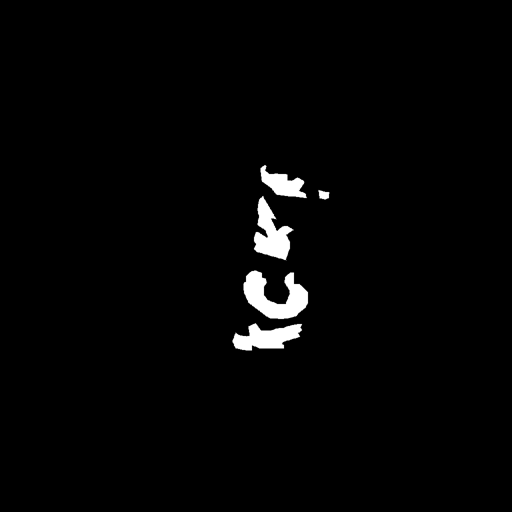} & \includegraphics[width=0.2\textwidth]{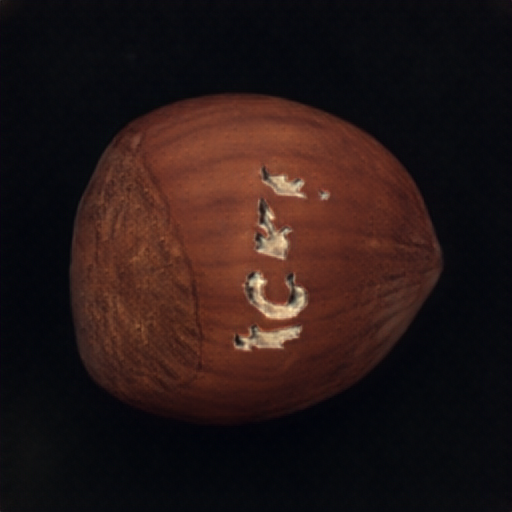} & \includegraphics[width=0.2\textwidth]{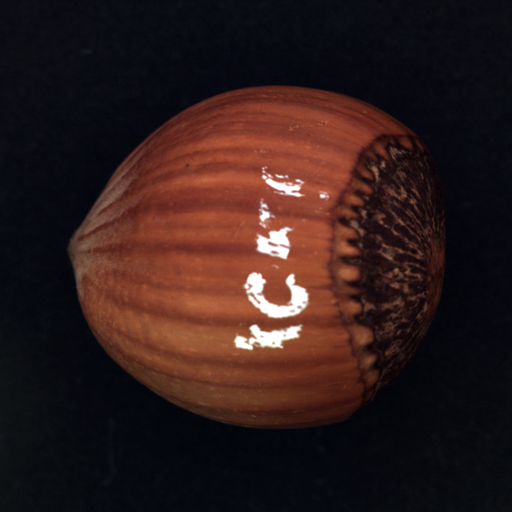} \\
\includegraphics[width=0.2\textwidth]{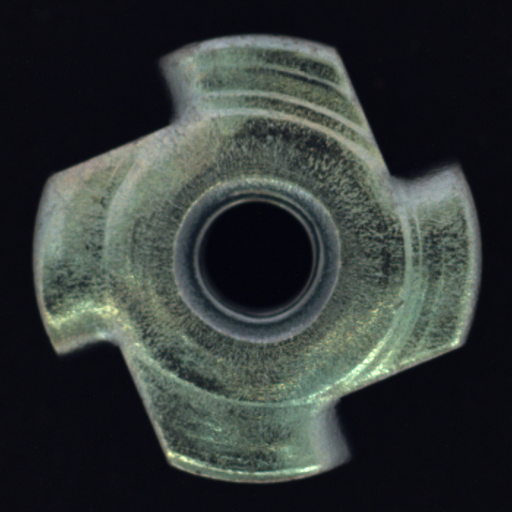} & \includegraphics[width=0.2\textwidth]{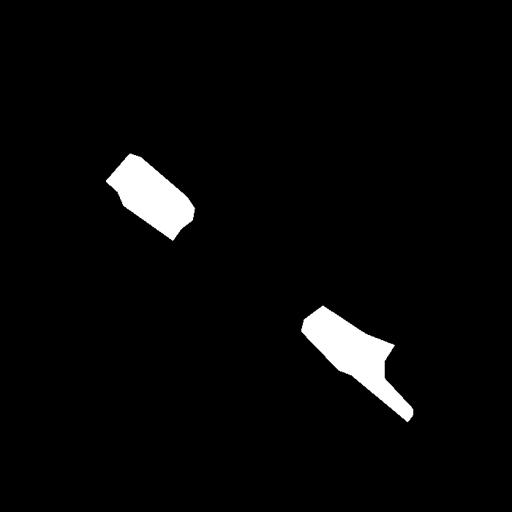} & \includegraphics[width=0.2\textwidth]{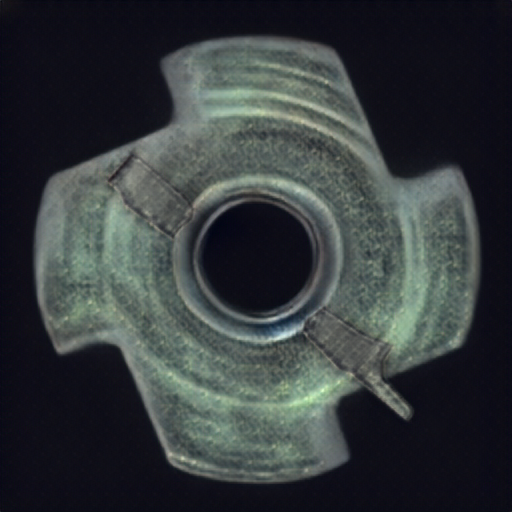} & \includegraphics[width=0.2\textwidth]{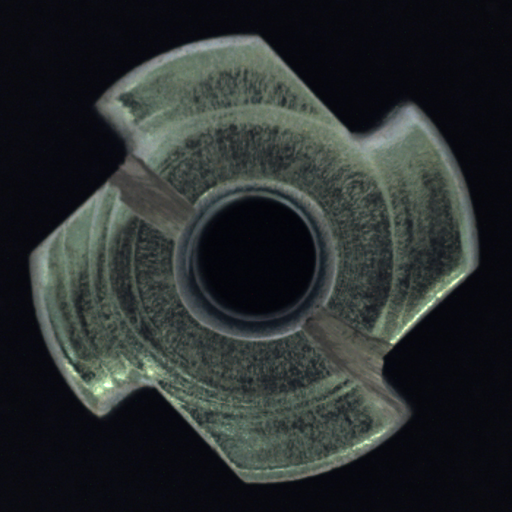} \\
\includegraphics[width=0.2\textwidth]{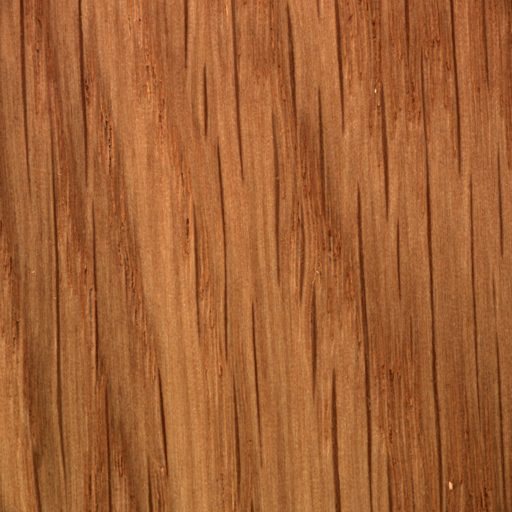} & \includegraphics[width=0.2\textwidth]{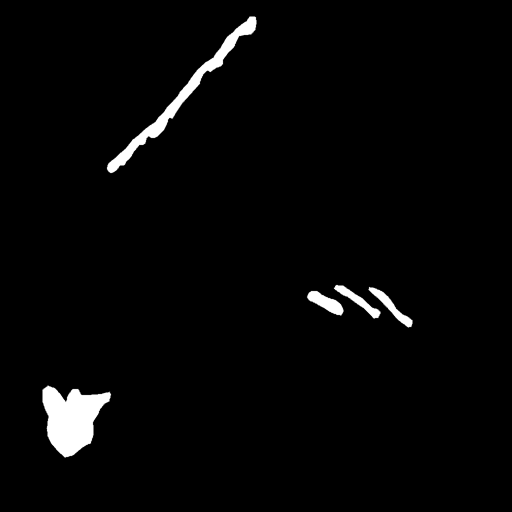} & \includegraphics[width=0.2\textwidth]{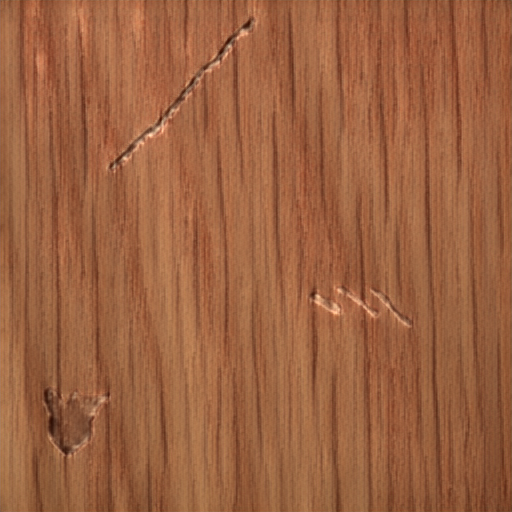} & \includegraphics[width=0.2\textwidth]{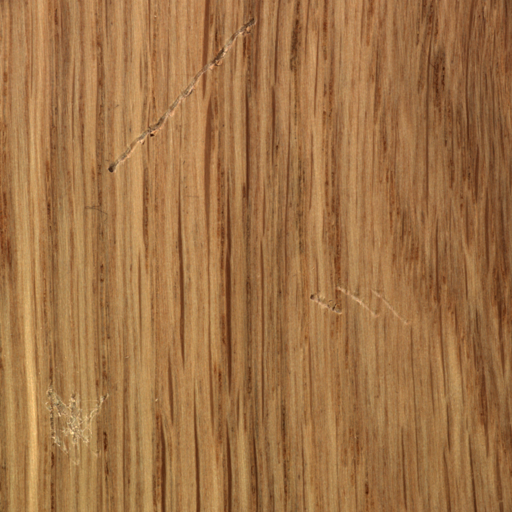} \\
\hline
\end{tabular}
\caption{UNPAIRED: More examples with different datasets with different prompts.}
\label{tab:training_unpaired_res_2}
\end{table}

\begin{table}[H]
\centering
\begin{tabular}{cccc}
\hline
Input Image   & Mask & Model Output & Ground truth \\

 \includegraphics[width=0.2\textwidth]{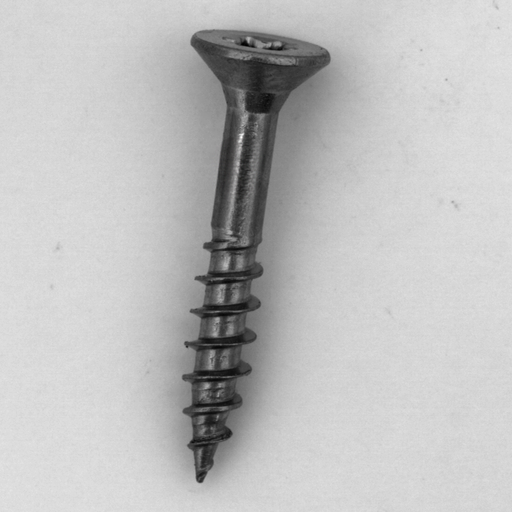} & \includegraphics[width=0.2\textwidth]{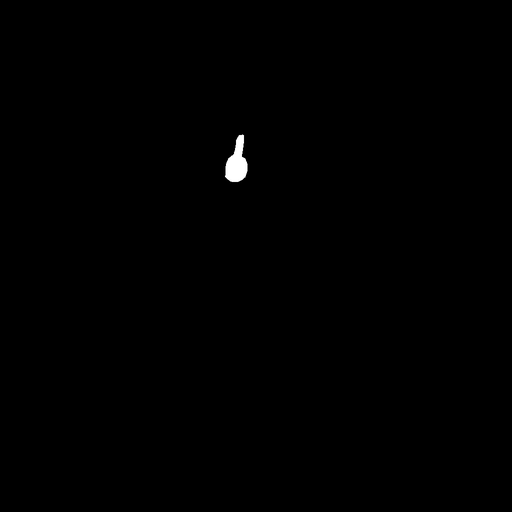} & \includegraphics[width=0.2\textwidth]{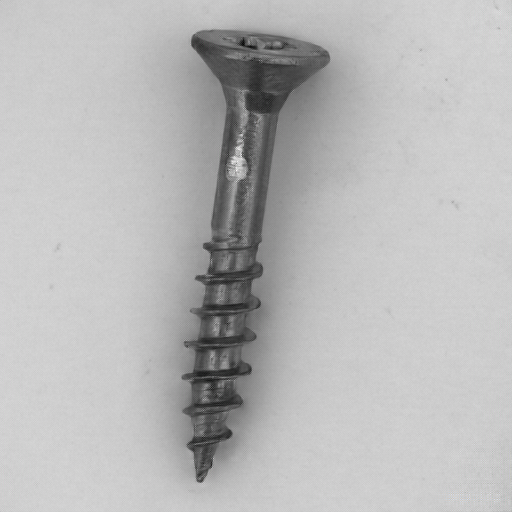} & \includegraphics[width=0.2\textwidth]{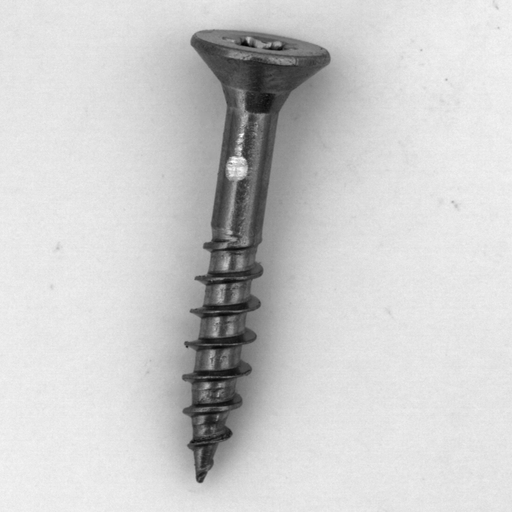} \\
\hline

\end{tabular}
\caption{Comparison of model output against ground truth for screw images. The images show the mask used, the input image, the model's generated output, and the actual ground truth image. This demonstrates the model's ability to generate defects accurately after 1,000 training steps with a small paired dataset of 15 images with the prompt "add scratch".}
\label{tab:screw_table3}

\end{table}

\begin{table}[H]
\centering
\begin{tabular}{cccc}
\hline
Input Image & Source & Model Output & Ground truth \\ 
\includegraphics[width=0.2\textwidth]{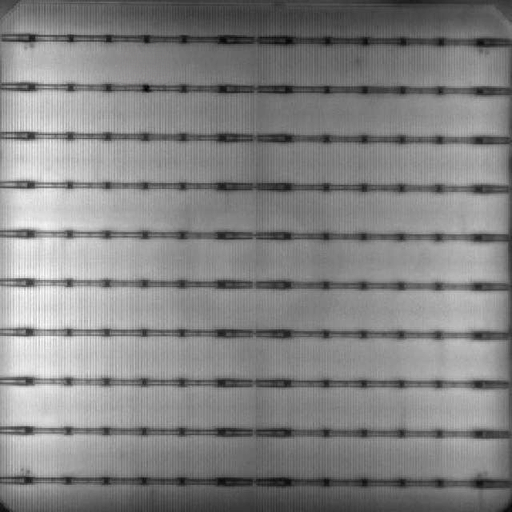} & \includegraphics[width=0.2\textwidth]{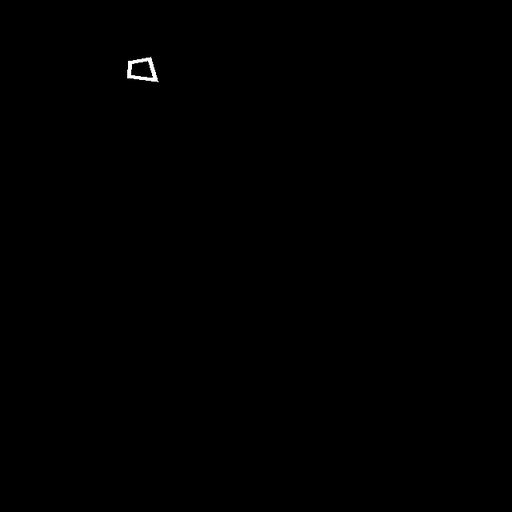} & \includegraphics[width=0.2\textwidth]{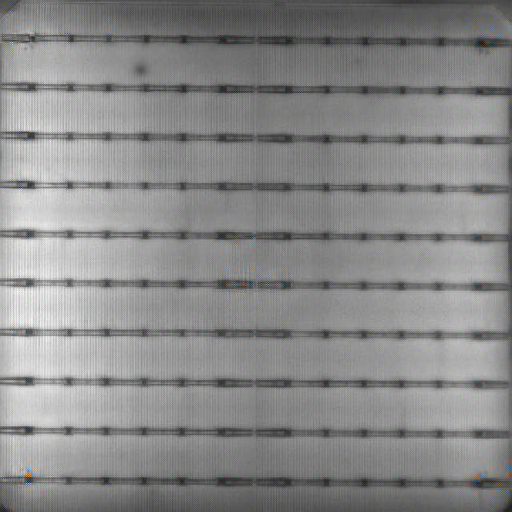} & \includegraphics[width=0.2\textwidth]{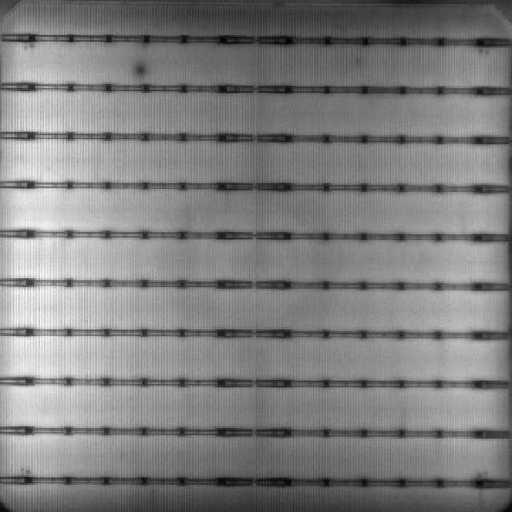} \\ 
\includegraphics[width=0.2\textwidth]{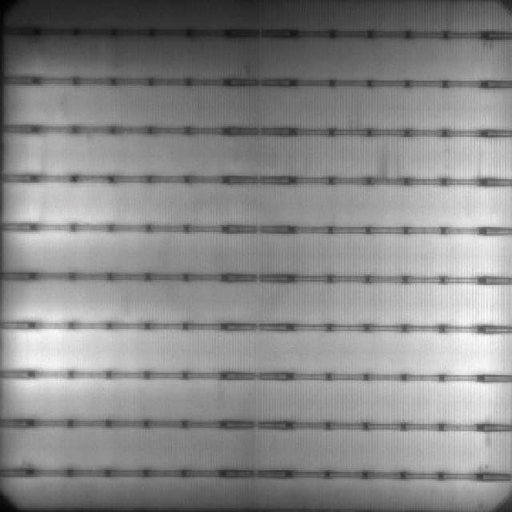} & \includegraphics[width=0.2\textwidth]{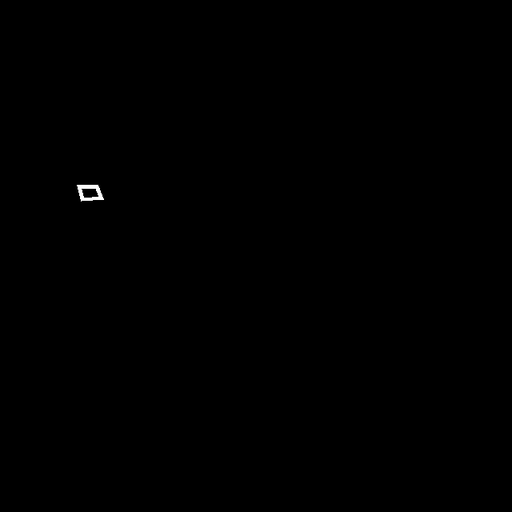} & \includegraphics[width=0.2\textwidth]{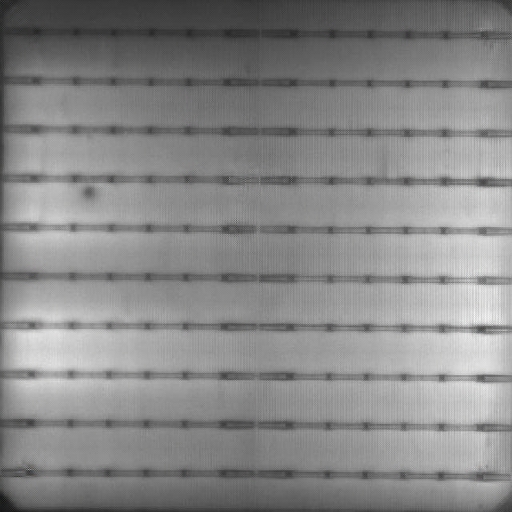} & \includegraphics[width=0.2\textwidth]{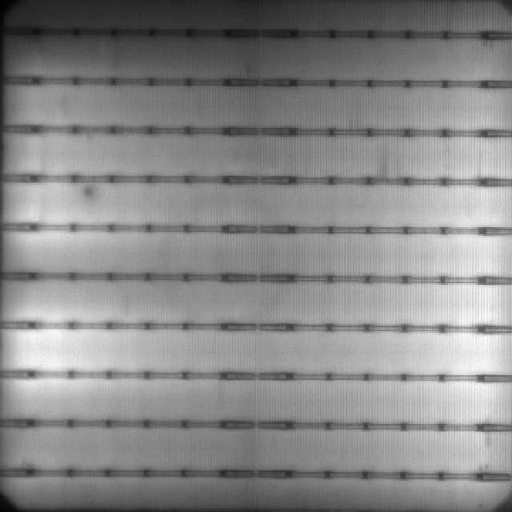} \\ 
\includegraphics[width=0.2\textwidth]{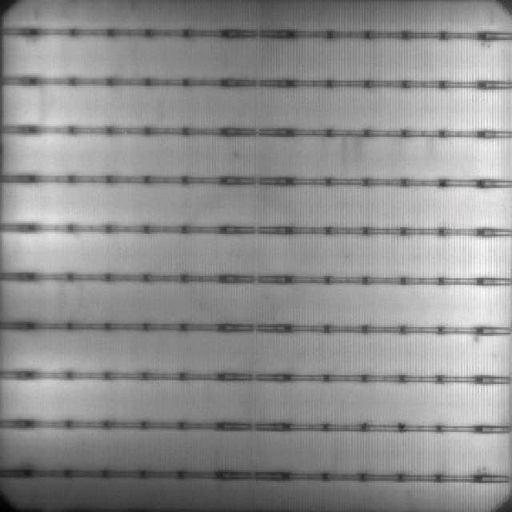} & \includegraphics[width=0.2\textwidth]{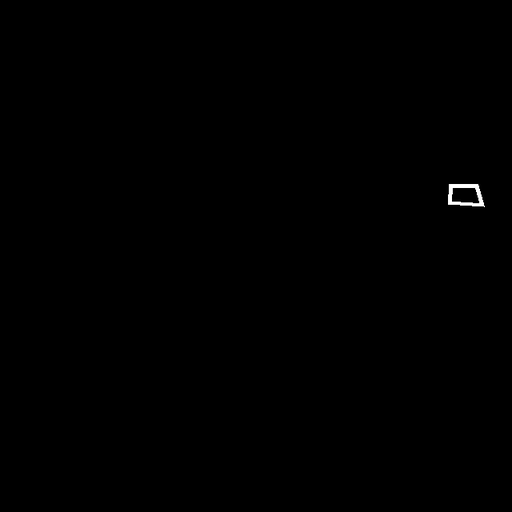} & \includegraphics[width=0.2\textwidth]{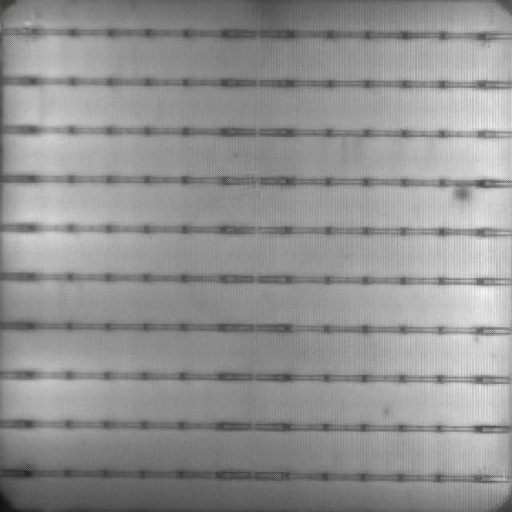} & \includegraphics[width=0.2\textwidth]{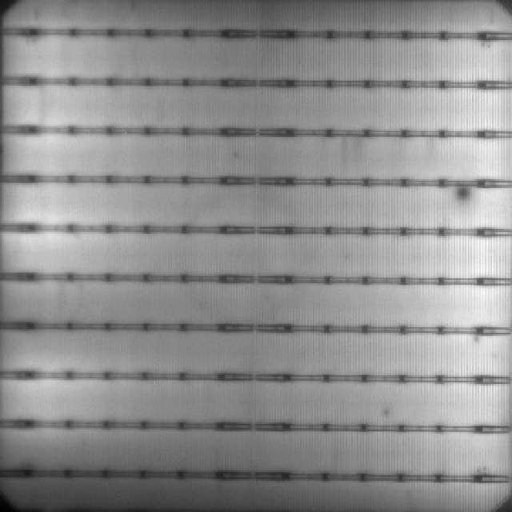} \\ \hline
\end{tabular}
\caption{Images generated by the industrial paired dataset with a different type of masks after 30 minutes of training with the prompt "add burn".}
\label{tab:industrial_samples_more}
\end{table}

\begin{table}[H]
    \centering
    \renewcommand{\arraystretch}{1.1}
    \setlength{\tabcolsep}{5pt}
    \begin{small}
    \begin{tabular}{@{}>{\centering\arraybackslash}m{0.26\textwidth} >{\centering\arraybackslash}m{0.16\textwidth} >{\centering\arraybackslash}m{0.16\textwidth} >{\centering\arraybackslash}m{0.16\textwidth} >{\centering\arraybackslash}m{0.16\textwidth}@{}}
        \toprule
        \textbf{Name} & \textbf{D\_S} & \textbf{D\_S\_AUG} & \textbf{CAS} & \textbf{NAS} \\ \midrule
        train/box\_loss        & 18.523 & 16.212 & 13.633 & \textbf{11.881} \\ \hdashline
        train/seg\_loss        & 17.821 & 16.124 & 0.704 & \textbf{0.6593} \\ \hdashline
        train/cls\_loss        & 18.076 & 17.421 & 0.656 & \textbf{0.601} \\ \hdashline
        train/dfl\_loss        & 12.277 & 12.812 & 11.483 & \textbf{10.729} \\ \bottomrule
    \end{tabular}
    \end{small}
    \caption{Training loss results for different experimental setups on the \textbf{Industrial panels} paired dataset. 
    \textbf{D\_S}: A simple split of the dataset into 70\% training and 30\% testing. 
    \textbf{D\_S\_AUG}: Similar to Data\_Split but with simple data augmentation applied. 
    \textbf{CAS} and \textbf{NAS}: Only images generated by our model were used.
    All models were validated on the same 30\% split of the data. Augmentation was performed only on the 70\% training data. A total of 19 images were used.}
    \label{tab:training_loss_panels}
\end{table}

\begin{table}[H]
    \centering
    \renewcommand{\arraystretch}{1.1}
    \setlength{\tabcolsep}{5pt}
    \begin{small}
    \begin{tabular}{@{}>{\centering\arraybackslash}m{0.26\textwidth} >{\centering\arraybackslash}m{0.16\textwidth} >{\centering\arraybackslash}m{0.16\textwidth} >{\centering\arraybackslash}m{0.16\textwidth} >{\centering\arraybackslash}m{0.16\textwidth}@{}}
        \toprule
        \textbf{Name} & \textbf{D\_S} & \textbf{D\_S\_AUG} & \textbf{CAS} & \textbf{NAS} \\ \midrule
        train/box\_loss        & 22.022 & 22.232 & 0.576 & \textbf{0.508} \\ \hdashline
        train/seg\_loss        & 21.458 & 20.793 & 0.714 & \textbf{0.650} \\ \hdashline
        train/cls\_loss        & 30.256 & 26.364 & 0.510 & \textbf{0.427} \\ \hdashline
        train/dfl\_loss        & 27.839 & 30.748 & 11.483 & \textbf{10.925} \\ \bottomrule
    \end{tabular}
    \end{small}
    \caption{Training loss results for different experimental setups on the \textbf{Tile} dataset. 
    \textbf{D\_S}: A simple split of the dataset into 70\% training and 30\% testing. 
    \textbf{D\_S\_AUG}: Similar to Data\_Split but with simple data augmentation applied. 
    \textbf{CAS} and \textbf{NAS}: Only images generated by our model were used. 
    All models were validated on the same 30\% split of the data. Augmentation was performed only on the 70\% training data. A total of 19 images were used.}
    \label{tab:training_loss}
\end{table}

\begin{table}[H]
    \centering
    \renewcommand{\arraystretch}{1.1}
    \setlength{\tabcolsep}{5pt}
    \begin{small}
    \begin{tabular}{@{}>{\centering\arraybackslash}m{0.5\textwidth} >{\centering\arraybackslash}m{0.5\textwidth}@{}}
        \toprule
        \textbf{Params} & \textbf{Values} \\ \midrule
        GAN Disc Type & vagan\_clip \\ \hdashline
        GAN Loss Type & multilevel\_sigmoid\_s \\ \hdashline
        Lambda GAN & 2.5 \\ \hdashline
        Lambda LPIPS & 10.0 \\ \hdashline
        Lambda L2 & 10.0 \\ \hdashline
        Lambda CLIPSim & 5.0 \\ \hdashline
        Dataset Folder & Required \\ \hdashline
        Train Image Prep & resized\_crop\_512 \\ \hdashline
        Test Image Prep & resized\_crop\_512 \\ \hdashline
        Eval Frequency & 100 \\ \hdashline
        Number of Samples for Evaluation & 100 \\ \hdashline
        Visualization Frequency & 100 \\ \hdashline
        Lora Rank for UNet & 8 \\ \hdashline
        Lora Rank for VAE & 4 \\ \hdashline
        Batch Size (Train) & 1 \\ \hdashline
        Number of Training Epochs & 10000 \\ \hdashline
        Max Training Steps & 10000 \\ \hdashline
        Checkpointing Steps & 500 \\ \hdashline
        Gradient Accumulation Steps & 1 \\ \hdashline
        Learning Rate & 5e-4 \\ \hdashline
        Learning Rate Scheduler & constant \\ \hdashline
        Warmup Steps & 500 \\ \hdashline
        Number of Cycles for LR & 1 \\ \hdashline
        Adam Beta1 & 0.9 \\ \hdashline
        Adam Beta2 & 0.999 \\ \hdashline
        Adam Weight Decay & 1e-2 \\ \hdashline
        Max Gradient Norm & 1.0 \\ \hdashline
        Use TF32 & False \\ \hdashline
        Mixed Precision & fp16 \\ \hdashline
        Enable Xformers & True \\ \bottomrule
    \end{tabular}
    \end{small}
    \caption{Hyperparameters used for training}
    \label{tab:hyperparameters}
\end{table}

\end{document}